\documentclass[letterpaper]{article} 
\usepackage{aaai2026}  
\usepackage{times}  
\usepackage{helvet}  
\usepackage{courier}  
\usepackage[hyphens]{url}  
\usepackage{graphicx} 
\usepackage{natbib}  
\usepackage{caption} 
\usepackage{algorithm}
\usepackage{algorithmic}
\usepackage{amsmath}
\usepackage{amsthm}
\usepackage{amsfonts}
\usepackage{amssymb}
\usepackage{subfigure}
\usepackage{booktabs}
\usepackage{multirow}
 
\theoremstyle{plain} 
\newtheorem{theorem}{Theorem}
\usepackage{newfloat}
\usepackage{listings}
\DeclareCaptionStyle{ruled}{labelfont=normalfont,labelsep=colon,strut=off} 
\floatstyle{ruled}
\newfloat{listing}{tb}{lst}{}
\floatname{listing}{Listing}
\makeatletter
\renewcommand{\@fnsymbol}[1]{\ensuremath{\dag}}
\makeatother

\title{Conditional Diffusion Model for Multi-Agent Dynamic Task Decomposition}
\author{
    Yanda Zhu\textsuperscript{\rm 1}, 
    Yuanyang Zhu\textsuperscript{\rm 2}\thanks{Corresponding author: Yuanyang Zhu and Caihua Chen.}, 
    Daoyi Dong\textsuperscript{\rm 3},
    Caihua Chen\textsuperscript{\rm 1}$^\dag$,
    Chunlin Chen\textsuperscript{\rm 4,1}
}
\affiliations{
    \textsuperscript{\rm 1} School of Management and Engineering, Nanjing University,
    \textsuperscript{\rm 2} School of Information Management, Nanjing University\\
    \textsuperscript{\rm 3} Australian Artificial Intelligence Institute, University of Technology Sydney,\\
    \textsuperscript{\rm 4} School of Robotics and Automation, Nanjing University\\
    {yandazhu@smail.nju.edu.cn}, {\{yuanyangzhu, chchen, clchen\}@nju.edu.cn}, {daoyidong@gmail.com}
}

\usepackage{bibentry}

\newcounter{appendixctr}
\makeatletter
\newcommand{\appsection}[1]{%
  \refstepcounter{appendixctr}
  \protected@edef\@currentlabel{\Alph{appendixctr}}  
  \section*{\Alph{appendixctr}.~#1}
  \addcontentsline{toc}{section}{\Alph{appendixctr}.~#1}
}
\makeatother

\begin{document}

\maketitle

\begin{abstract}
Task decomposition has shown promise in complex cooperative multi-agent reinforcement learning (MARL) tasks, which enables efficient hierarchical learning for long-horizon tasks in dynamic and uncertain environments.
However, learning dynamic task decomposition from scratch generally requires a large number of training samples, especially exploring the large joint action space under partial observability.
In this paper, we present the Conditional Diffusion Model for Dynamic Task Decomposition (C$\text{D}^\text{3}$T), a novel two-level hierarchical MARL framework designed to automatically infer subtask and coordination patterns.
The high-level policy learns subtask representation to generate a subtask selection strategy based on subtask effects.
To capture the effects of subtasks on the environment, C$\text{D}^\text{3}$T predicts the next observation and reward using a conditional diffusion model.
At the low level, agents collaboratively learn and share specialized skills within their assigned subtasks.
Moreover, the learned subtask representation is also used as additional semantic information in a multi-head attention mixing network to enhance value decomposition and provide an efficient reasoning bridge between individual and joint value functions.
Experimental results on various benchmarks demonstrate that C$\text{D}^\text{3}$T achieves better performance than existing baselines.

\end{abstract}

\section{Introduction}\label{introduction}
Cooperative multi-agent reinforcement learning (MARL) has achieved great improvements and holds great promise for real-world challenging problems, such as sensor networks~\cite{zhang2011coordinated}, coordination of robot swarms~\cite{huttenrauch2017guided}, and autonomous vehicles~\cite{pham2018cooperative}.
Learning effective control policies under partial observation for coordinating such systems remains challenging.
The centralized training with decentralized execution (CTDE) paradigm~\cite{oliehoek2008optimal, kraemer2016multi} alleviates partial observability yet struggles with the exponential growth of the joint action-observation space as agent numbers increase, which makes exploration of valuable states rare and coordination difficult.

To deal with uncertainty and adapt to the dynamics of an environment, all agents learn and share a decentralized policy network under the CTDE framework. 
Memory-based architectures, such as recurrent neural networks (RNNs), long short-term memory (LSTM)~\cite{hochreiter1997long}, and gated recurrent units (GRUs), help agents to capture long-term dependencies in their action-observation history~\cite{vdn,qmix,qtran,qplex,liu2025mixrts}.
Recent transformer-based methods have shown superior performance by modeling both long- and short-term dependencies, offering a powerful solution to partial observability~\cite{parisotto2020stabilizing, yang2022transformer, mat}. 
However, parameter sharing among agents can lead to similar behavior, hindering diversity.
The challenge is to balance agent specialization and dynamic sharing to promote cooperation~\cite{christianos2021scaling}.

A natural solution to this challenge is decomposing complex tasks into subtasks~\cite{butler2011condensed}.
This decomposition not only simplifies the task but also allows agents to focus on solving specific subtasks, which can reduce the complexity of the action-observation space and enhance overall learning efficiency.
Building on this idea, recent research has explored the integration of roles and skills into MARL.
In the learning of roles~\cite{wang2020roma, wang2021rode, li2021celebrating}, skills~\cite{yang2022ldsa, liu2022heterogeneous}, or groups~\cite{zang2023automatic}, existing works generally use a simple network structure to extract action representations for agents and may neglect fully considering the dynamic interactions among agents and the environment.
The representational capacity of such a setting poses a bottleneck when trying to learn distinct latent representations for all subtasks.

Diffusion Models~\cite{ho2020denoising,sohl2015deep}, a novel class of generative models known for their impressive performance in image generation tasks, offer a promising avenue to address such challenges.
These models are well-suited for handling the stochasticity inherent in complex environments due to their ability to model stochastic processes through iterative denoising. 
Furthermore, since the spaces of histories and states in MARL are often continuous and high-dimensional, diffusion models are particularly effective because of their robust representational capacity in expansive spaces.
These benefits of the diffusion model stimulate our thinking in MARL domains, i.e., \textit{can we harness modern generative models, such as diffusion models, trained on offline data and capture useful latent representation that facilitates online MARL?}

To explore this, we propose the Conditional Diffusion Model for Dynamic Task Decomposition C$\text{D}^\text{3}$T.
To reflect the potential characters of agents and subtasks, a set of action representations is used as the input conditioned on observations and actions of agents, while the reward of the environment and the next observations are used as the output.
With this representation, we can derive subtasks by clustering and devise a subtask selection mechanism that assigns an agent to a subtask. 
Due to characters encoded in the subtask representation, this mechanism could select and share proper skills for agents based on parameter sharing.
Benefit from the powerful representational capacity of the diffusion model, C$\text{D}^\text{3}$T not only owns a better ability to model stochastic processes through the iterative denoising inductive bias but also learns distinguishable subtasks to explore the environment.
Inspired by recent works addressing spurious correlations between global states and joint values~\cite{li2022deconfounded,wang2023asn,liu23be}, we incorporate subtask representations with global state information to better estimate credit assignment.

We evaluate C$\text{D}^\text{3}$T across a range of benchmarks, including Level-based Foraging (LBF), StarCraft Multi-Agent Challenge (SMAC)~\cite{samvelyan2019starcraft}, and SMACv2~\cite{ellis2023smacv2}.
The results show that our C$\text{D}^\text{3}$T improves the performance on SMAC compared to the baselines, especially on \textit{Hard} and \textit{Super Hard} scenarios.
Ablation studies confirm the efficacy of task decomposition and credit assignment design, and visualizations illustrate meaningful dynamic task decompositions and cooperation.

\begin{figure*}[h]
\begin{center}
    \includegraphics[width=0.95\textwidth]{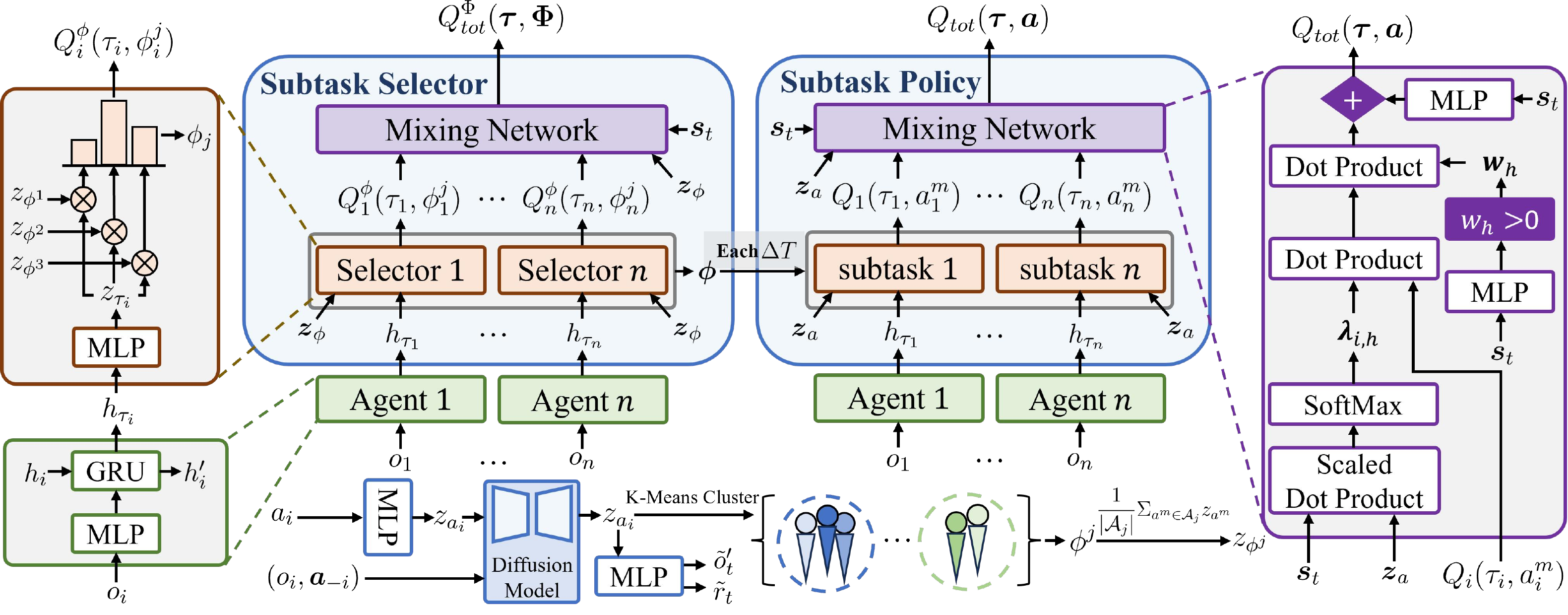}
\end{center}
\vspace{-0.5cm}
\caption{
The overall framework of C$\text{D}^\text{3}$T. 
We first derive a latent action representation $z_{a_i}$ for each agent from its action space, conditioned on its local observation $o_i$ and other agents' one-hot actions $\boldsymbol{a}_{-i}$, to pretrain a diffusion model.
Latent representations are then clustered to define subtask-specific action spaces. 
The subtask selector and subtask policy share the same architecture with different parameters.
At every $\Delta T$ steps, the selector assigns a subtask to each agent and estimates the joint Q-value $Q_{tot}^\Phi$ using the global state $s_t$ and subtask representation $\boldsymbol{z}_\phi$, while the subtask policy computes $Q_{tot}$ with $s_t$ and the action representation $\boldsymbol{z}_a$.
}
\vspace{-0.5cm}
\label{frame}
\end{figure*}

\section{Preliminaries}
Our work focuses on a fully cooperative multi-agent task with only partial observation for each agent, which typically is modeled as a decentralized partially observable Markov decision process (Dec-POMDP)~\cite{oliehoek2016concise} and described with the tuple $\mathcal{M}=\left\langle \mathcal{I}, \mathcal{S}, \mathcal{A}, P, R, \Omega, O, \gamma \right\rangle$.
At each time step, each agent $i \in \mathcal{I} $ receives an observation $o_i \in \Omega$, drawn from the observation function $O(s, i)$, where $s \in \mathcal{S}$ is the global state of the environment, and selects an action $a_i \in \mathcal{A}$, producing a joint action $\boldsymbol{a} = (a_1, \dots, a_n)$.
This joint action would lead to the next state $s'$ according to the state transition function $P(s'|s)$, and all agents would receive a shared team reward $r = R(s, \boldsymbol{a})$.
We use $\boldsymbol{\tau} \in \mathcal{T}\equiv(\Omega \times \mathcal{A})^*$ to denote the joint action-observation history, where $\boldsymbol{\tau} = (\tau_1, \dots, \tau_n)$, and $\tau_i = (o_i^1, a_i^1, \dots, o_i^{t-1}, a_i^{t-1}, o_i^t)$ represents the trajectory of agent $i$.
The target is to find the optimal joint policy $\boldsymbol{\pi}(\boldsymbol{a}|\boldsymbol{\tau})$ that maximizes the discounted return, defined as $Q^\pi(\boldsymbol{\tau}, \boldsymbol{a}) = \mathbb{E}\left[\sum_{t=0}^\infty \gamma^t R(s_t, \boldsymbol{a}_t) \mid s_0 = s, \boldsymbol{a}_0 = \boldsymbol{a}, \boldsymbol{\pi}\right]$, where $\gamma \in [0, 1)$ is the discount factor.

Unlike recent MARL works~\cite{vdn,qmix,qtran,qplex}, we propose to decompose a fully cooperative multi-agent task into subtasks and present the definition of subtasks in the following.

\textbf{Definition 1} (Subtasks)\textbf{.} \textit{Given a cooperative task $\mathcal M=\left\langle \mathcal{I,S,A},P,R,\Omega,O,\gamma \right\rangle$, we assume there exists a set of $g$ subtasks, denoted as $\Phi=\{\phi^1,\phi^2,...,\phi^g\}$, where $g\in\mathbb{N}^+$ is unknown and considered as a tunable hyperparameter.
Each subtask is expressed as a tuple $\left\langle \mathcal{M}_{\phi^j},\pi_{\phi^j}\right\rangle$, where $j\in \{1,2,...,g\}$ is the identity of subtask, $\mathcal{M}_{\phi^j}=\left\langle \mathcal{I}_{\phi^j},\mathcal{S},\mathcal{A}_{\phi^j},P,R,\Omega_{\phi^j},O,\gamma\right\rangle$ and $\pi_{\phi^j}:\mathcal{T}\times\mathcal{A}_{\phi^j}\rightarrow[0,1]$.
$\mathcal{I}_{\phi^j}$ is the set of agents assigned to subtask $\phi^j$ and each agent can only select one subtask to solve at each timestep, i.e., $\mathcal I_{\phi^j} \subset \mathcal{I}$, $\cup_{\phi^j}\mathcal{I}_{\phi^j}=\mathcal{I}$ and $\mathcal{I}_{\phi^j}\cap\mathcal{I}_{\phi^k}=\varnothing$ if $j\neq k$.
$\mathcal{A}_{\phi^j}$ is the action space of the subtask, $\mathcal{A}_{\phi^j}\subset \mathcal{A}$, $\cup_{\phi^j}\mathcal{A}_{\phi^j}=\mathcal{A}$, but we allow action spaces of different subtasks to overlap: $\lvert \mathcal{A}_{\phi^j}\cap \mathcal{A}_{\phi^k}\rvert\geq0$ if $j\neq k$.
Each agent $i\in\mathcal{I}_{\phi^j}$ shares the policy parameters of $\pi_{\phi^j}$.} 

With the set of subtasks $\Phi$ defined, each agent $i_{\phi^j} \in \mathcal{I}_{\phi^j}$ is assigned subtask $\phi^j$ through a shared subtask selector.
This enables the learning of subtask-specific policies $\pi_{\phi^j}: \boldsymbol{\tau} \times \mathcal{A}_{\phi^j}$ for each subtask.
Our objective is to learn the optimal set of subtasks $\Phi^*$ that maximizes the expected global return $Q_{tot}^\Phi(\boldsymbol\tau,\boldsymbol a)=\mathbb{E}_{s_{1:\infty},\boldsymbol a_{1:\infty}} [\sum\limits_{t=0}^\infty\gamma^tr_t|s_0=s,\boldsymbol a_0=\boldsymbol a,\Phi ]$.

\section{Method}
Our solution for multi-agent dynamic task decomposition is illustrated in \textit{Fig.~\ref{frame}}.
We begin by describing how to construct action semantic representations that enable the decomposition of multi-agent tasks.
Next, we explain how the representations are leveraged to generate subtasks.
Based on the generated subtasks and their corresponding latent representations, we introduce a hierarchical architecture consisting of a subtask selector and a set of subtask policies.
Finally, we detail the training objective and inference strategy for both the subtask selector and the subtask policies.

\subsection{Action Representation Learning via Diffusion}
The latent action representations are designed to induce diverse subtasks with distinct responsibilities, capturing characteristic agent behaviors for more appropriate subtask selection.
While this design allows C$\text{D}^\text{3}$T to adapt to dynamic environments, it may lead to rapid subtask shifts and instability during learning. Moreover, if the induced subtasks are overly similar, decomposition becomes ineffective.
Therefore, two key challenges arise: 1) ensuring temporal stability to maintain adaptability, and 2) enhancing subtask diversity through efficient modeling.

To this end, we first construct the action encoder component to map the one-hot action $a_i$ of agent $i$ to the $d$-dimensional representation $z_{a_i}$, which serves as unmodified examples $z_0$.
Then the UNet backbone with cross-attention is employed as a flexible feature extractor in the denoising network $\epsilon_{\theta_d}(z_k,k,o_i,a_{-i})$ to recover $z_{a_i}$ from Gassian noise $\epsilon\sim\mathcal{N}({0},{I})$ conditioned on corresponding $o_i$ and $a_{-i}$.
Following the simplified objective~\cite{ho2020denoising}, we formulate a learning objective for action representations via the conditional diffusion model parameterized by $\theta_d$, which is trained by minimizing the loss function:
\begin{align}
\mathcal{L}_d(\theta_d) = \mathbb{E}_{\boldsymbol{\epsilon} \sim \mathcal{N}(\boldsymbol{0}, \boldsymbol{I}), (\boldsymbol{o}, \boldsymbol{a}) \sim \mathcal{D}} [ \left\Vert \boldsymbol{\epsilon} -\boldsymbol{\epsilon}_{\theta_d}  (\boldsymbol z_k, k, o_i, \boldsymbol{a}_{-i}  ) \right\Vert^2 ],
\end{align}
where $\mathcal{D}$ denotes the replay buffer, $k$ is the diffusion iteration uniformly sampled from $\{1, 2, \ldots, K\}$ and $z_k$ is the noisy version of $z_0$.
Here, $o_i$ denotes the observation of agent $i$, while $\boldsymbol a_{-i}$ denotes the joint actions of all other agents.
The detailed derivation can be found in Appendix~\ref{appendix:diffusion}.

Action semantics, where different actions have varying effects on other agents, should influence the environment or their private properties.
To further extract the influence of an action through the induced reward and the change in local observations, we leverage the action representations generated by the diffusion model to predict the next observations $o_i'$ and global reward $r$ based on $o_i$ and $\boldsymbol a_{-i}$.
The prediction objective can be further rewritten as a loss function:
\begin{align}\label{eq8}
\mathcal{L}_{p}\left(\theta\right) = 
&\; \mathbb{E}_{\left( \boldsymbol{o},\boldsymbol{a},r,\boldsymbol{o}' \right)\sim\mathcal{D}} \left[
    \sum_i \left\| f_{do}\left( z_{a_i},o_i,\boldsymbol{a}_{-i} \right) - o_i' \right\|_2^2 \right. \notag \\
&\quad \left. + \lambda_{dr} \sum_i \left( f_{dr} \left( z_{a_i},o_i,\boldsymbol{a}_{-i} \right) - r \right)^2 
\right],
\end{align}
where $f_{do}$ and $f_{dr}$ are predictors of observations and global rewards, respectively, and $\lambda_{dr}$ is a scaling factor.
Here, the summation covers all agents, and the whole action representation learning is parameterized by $\theta$.
Thus, the objective for action representation learning combines the prediction loss and the diffusion model loss with a scaling factor $\eta_d$:
\begin{equation}\label{eq:loss of prediction}
\begin{array}{r}
    \mathcal{L}(\theta)=\mathcal{L}_p({\theta})+\eta_d\mathcal{L}_d(\theta_d).
\end{array}
\end{equation}

The current formulation does not explicitly enforce subtask specialization, which is essential for ensuring behavioral diversity across different subtasks.
Prior approaches typically encourage specialization via explicit regularization techniques~\cite{christianos2021scaling,wang2020roma}.
In contrast, we utilize the diffusion model as a flexible feature extractor and enhance its UNet backbone with cross-attention to generate action representations that capture multimodal distributions.
This property naturally induces subtask diversity without additional regularization.

\subsection{Subtask Dynamic Decomposition}
The action representation plays a critical role in assigning agents to the most suitable subtasks, where agents dealing with the same subtask share their learning of specific abilities. 
Considering that an agent's subtask history reflects not only its behavioral history but is also task-independent, we perform $k$-means clustering over these action representations to decompose the task into a finite set of subtasks after sampling and learning for the initial 50$K$ timesteps of the overall training process.
Given the action representations of subtask $\phi^j$, the subtask representation $z_{\phi^j}$ can be derived as $z_{\phi^j} = \frac{1}{|\mathcal A_j|} \sum_{a^m \in \mathcal A_j} z_{a^m}$, where $\mathcal A_j$ represents the decomposed action space of the subtask $\phi^j$ and $a^m$ denotes actions in this restricted space.
With a hierarchical network consisting of the subtask selector and the subtask policies, subtask representations can be used to assign the most suitable subtask to an agent over a specific time interval $\Delta T$.
Similarly, an MLP layer and a shared GRU layer network are shared by all agents to obtain the historical information of the agent's local observations and actions, which is parameterized by $\theta_{\tau_\phi}$ and compiles it into a vector $h$ for our subtask selection.

When selecting subtasks every $\Delta T$ timesteps, the subtask selector encodes the historical information $h_{\tau}$ of each agent into the hidden layer variable ${\boldsymbol z}_{\tau}\in\mathbb{R}^d$ with a $\xi_\phi$-parameterized encoder $f_{\phi}\left( :;\xi_\phi \right)$.
Then it is used to estimate the expected return of agent $i$ in subtask $\phi^j$ with the hidden layer representation of each subtask space $ Q^{\phi}_{i}\left( \tau_i,\phi^j \right)={\boldsymbol z}^T_{\tau_i}{\boldsymbol z}_{\phi^j}.$
The subtask space corresponding to the maximum Q-value is assigned to each agent.
In the next $\Delta T$ time step, each agent learns its policy in the restricted subtask space.

\subsection{Learning Decomposition with Credit Assignment}
The global state $\boldsymbol{s}$ in multi-agent systems contains rich information, yet only a subset is relevant to individual decision-making. To extract meaningful abstractions without relying on expert knowledge, we follow~\cite{li2022deconfounded,liu23be,xuhigh} and mitigate spurious correlations between $\boldsymbol{s}$ and $Q_{tot}$ by leveraging agents’ local histories $\tau_i$ in partially observable settings.
To improve decomposition accuracy and credit assignment, we introduce an intervention-based adjustment function during training, which adheres to the Individual-Global-Maximum (IGM) principle in Appendix~\ref{appendix:credit assignment}.

Specifically, the credit $\lambda^{\phi}_{h,i}$ for subtask selector is computed with the subtask representation $\boldsymbol z_{\phi}$ and the global state $\boldsymbol{s}$ through a dot-product attention as
\begin{equation}
\begin{array}{r}
    \lambda^\phi_{h,i} = \frac{\exp\left((\boldsymbol{w}_{\boldsymbol z_{\phi}} \boldsymbol{z}_{\phi})^\top \operatorname{ReLU}(\boldsymbol{w}_s \boldsymbol{s})\right)}{\sum_{i=1}^N \exp\left((\boldsymbol{w}_{\boldsymbol z_{\phi}} \boldsymbol{z}_{\phi})^\top \operatorname{ReLU}(\boldsymbol{w}_s \boldsymbol{s})\right)},
\end{array}
\end{equation}
where $\boldsymbol{w}_{\boldsymbol s}$, $\boldsymbol{w}_{\boldsymbol z_{\phi}}$ are learnable weight matrices, and ReLU is the element-wise rectified-linear activation. $\lambda_{h,i}$ is positive with softmax operation to ensure monotonicity and $h$ is the number of attention heads.
The softmax ensures each $\lambda^{\phi}_{h,i}$ is positive and that $\sum_{i}\lambda^{\phi}_{h,i}=1$.
Then the joint action function $Q_{tot}^{\Phi}$ of the subtask selector can be estimated as 
\begin{equation}\label{eq: update subtask selector}
\begin{array}{r}
    Q_{tot}^{\Phi}=c_\phi(s)+\sum \limits_{h=1}^{H}w_h^\phi\sum \limits_{i=1}^{N}\lambda_{h,i}^\phi Q_i^{\phi^j}(\tau_i,\phi^j_i),
\end{array}
\end{equation}
where $c_\phi(s)$ is learned by a neural network with the global state $s$ as the input.
The joint action function $Q_{tot}^{\Phi}$ of the subtask selector can be optimized by minimizing TD loss:
\begin{align}
\mathcal{L}_{ss}( \theta_{\tau_\phi}, \xi_\phi ) 
= &\; \mathbb{E}_\mathcal{D} \Big[ \Big( 
    \sum\limits_{\Delta t=0}^{\Delta T-1} r_{t+\Delta t} 
    + \gamma \max\limits_{\boldsymbol\Phi'} \bar{Q}_{tot}^{\Phi}( s_{t+\Delta T}, \boldsymbol{\Phi}' ) \notag \\
&\quad - Q_{tot}^{\Phi}( s_t, \boldsymbol{\Phi}_t ) 
\Big)^2 \Big],
\end{align}
where $\xi_\phi$ denotes the parameters of the mixing network, $\boldsymbol{\Phi}=\left\langle \phi^1, \phi^2, \ldots, \phi^N \right\rangle$ is the joint subtask of all agents, and the expectation is taken over mini-batches sampled uniformly from the replay buffer $\mathcal{D}$.

During the timestep of $\Delta T$, each agent follows the subtask assigned by the high-level selector.
When the agent is assigned to the corresponding subtask, it selects its action in the action space of the subtask.
Therefore, each subtask has a corresponding policy $\pi_{\phi^j}:\mathcal{T}\times \mathcal A_{\phi^j} \rightarrow\left[ 0,1 \right]$, which is defined in the restricted subtask action space and updates such a policy network. 
To take full advantage of the action representation of the corresponding subtask space, we also use the mechanism to compute the joint function $Q_{tot}$. 
Here, we use the shared MLP layer and GRU layer parameterized by $\theta_{\tau}$ in the same way to compile the local action-observation information history $\tau$ into a vector $h_\tau$.
For each subtask policy, we use a fully connected network $f_{\phi^{'}}\left( h_\tau;\zeta_{\phi} \right)$ parameterized by $\zeta_{\phi}$ to represent it.
Thus we estimate the individual value of agent $i$ by choosing a primitive action $a^m_i$ as $Q_i\left( \tau_i,a^m_i \right)=\boldsymbol z_{\tau_i}^{T}\boldsymbol z_{a^m_i}$.

Similar to the value function factorization of the subtask selector, the action representation $\boldsymbol{z}_a$ and the global state $\boldsymbol{s}$ are still fed into the intervention function to estimate credits.
Here, the action representations are restricted in the action space of the subtasks assigned to the agent, rather than the subtask representations in the subtask selector, so the credit $\lambda_{h,i}$ for the subtask policy is computed as
\begin{equation}
\begin{array}{r}
	    \lambda_{h,i} = \frac{\exp\left((\boldsymbol{w}_{z_{a}} \boldsymbol{z}_{a})^\top \operatorname{ReLU}(\boldsymbol{w}_s \boldsymbol{s})\right)}{\sum_{i=1}^N\exp\left((\boldsymbol{w}_{z_a} \boldsymbol{z}_{a})^\top \operatorname{ReLU}(\boldsymbol{w}_s \boldsymbol{s})\right)},
\end{array}
\end{equation}
where $\boldsymbol{w}_{ s}$, $\boldsymbol{w}_{z_a}$ are the learnable parameters, and ReLU is the activation function. 
$\lambda_{h,i}$ is positive with softmax operation to ensure monotonicity and $h$ is the number of attention heads.
The joint value function of the subtask policy is predicted based on the credits and factorized Q-values 
\begin{equation}\label{eq: update subtask poicy}
\begin{array}{r}
    Q_{tot}=c(s)+\sum \limits_{h=1}^{H}w_{h}\sum \limits_{i=1}^{N}\lambda_{h,i} Q_i(\tau_i,a_i^m),
\end{array}
\end{equation}
where $c(s)$ is learned by a neural network with the global state $s$ as the input.
The formulation gives the TD loss for subtask policies:
\begin{align}
    \mathcal{L}_{s}( \theta_{\tau}, \xi) = \mathbb{E}_\mathcal{D} [ ( r + \gamma \max\limits_{\boldsymbol a'}\bar{Q}_{tot}( s', \boldsymbol{a}' )
     - Q_{tot}( s, \boldsymbol{a} ) )^2 ],
\end{align}
where the parameters of the mixing network are denoted by $\xi$ and $\bar{Q}_{tot}$ is a target network.
Training samples are drawn uniformly from the same replay buffer~$\mathcal{D}$ used for the high-level selector.
Under the CTDE paradigm, the selector, subtask policies, and individual utility networks are used jointly at execution time, but only local information is required for each agent to act.
Pseudocode for the complete C$\text{D}^\text{3}$T algorithm is given in Appendix~\ref{app:pseudo code}.

\begin{figure}[h]
  \centering
  \includegraphics[width=\linewidth]{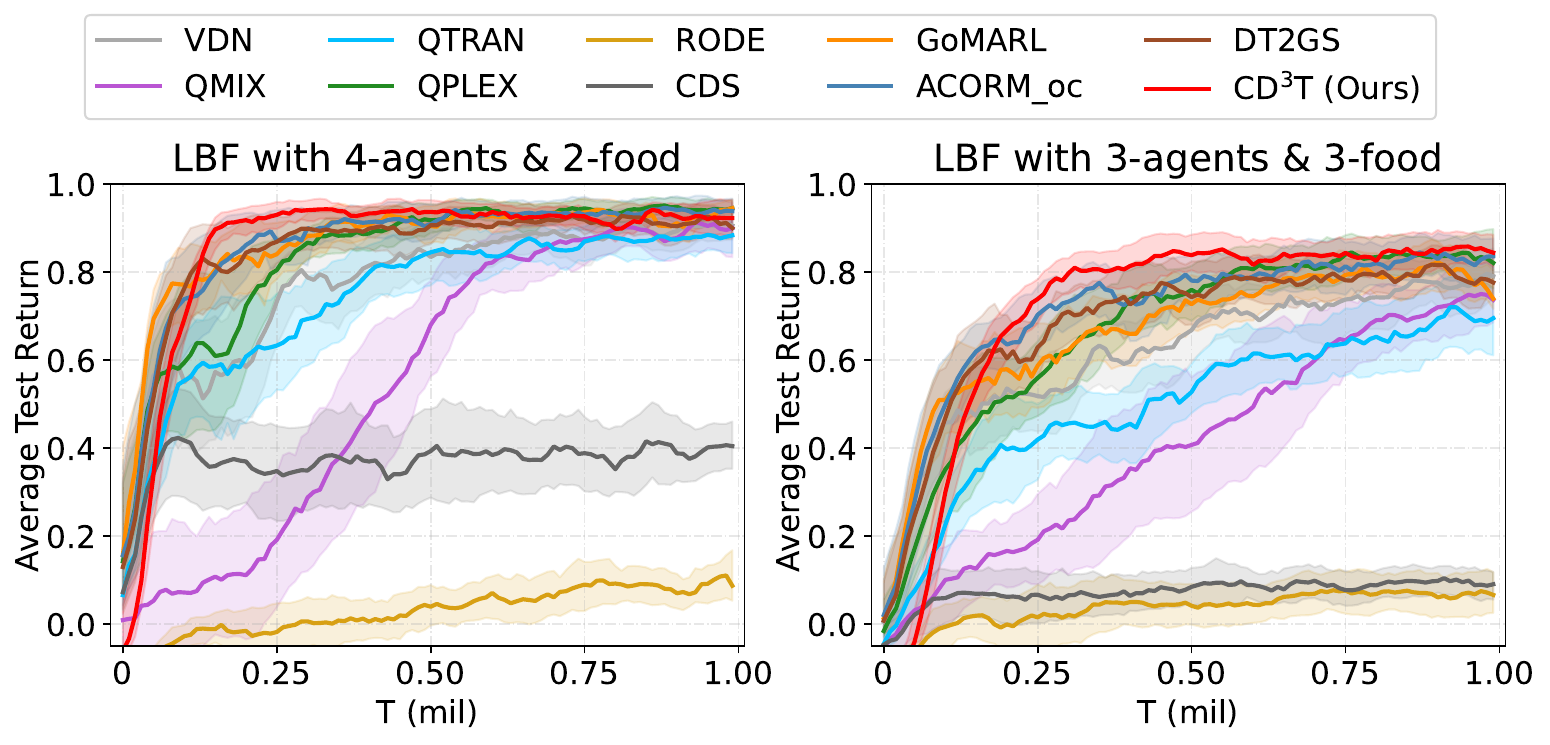}
  \vskip -0.05in
  \caption{Performance comparison with baselines on LBF.}
  \vspace{-0.3cm}
  \label{fig:lbf}
\end{figure}

\begin{figure*}[h]
\begin{center}
    \includegraphics[width=\linewidth]{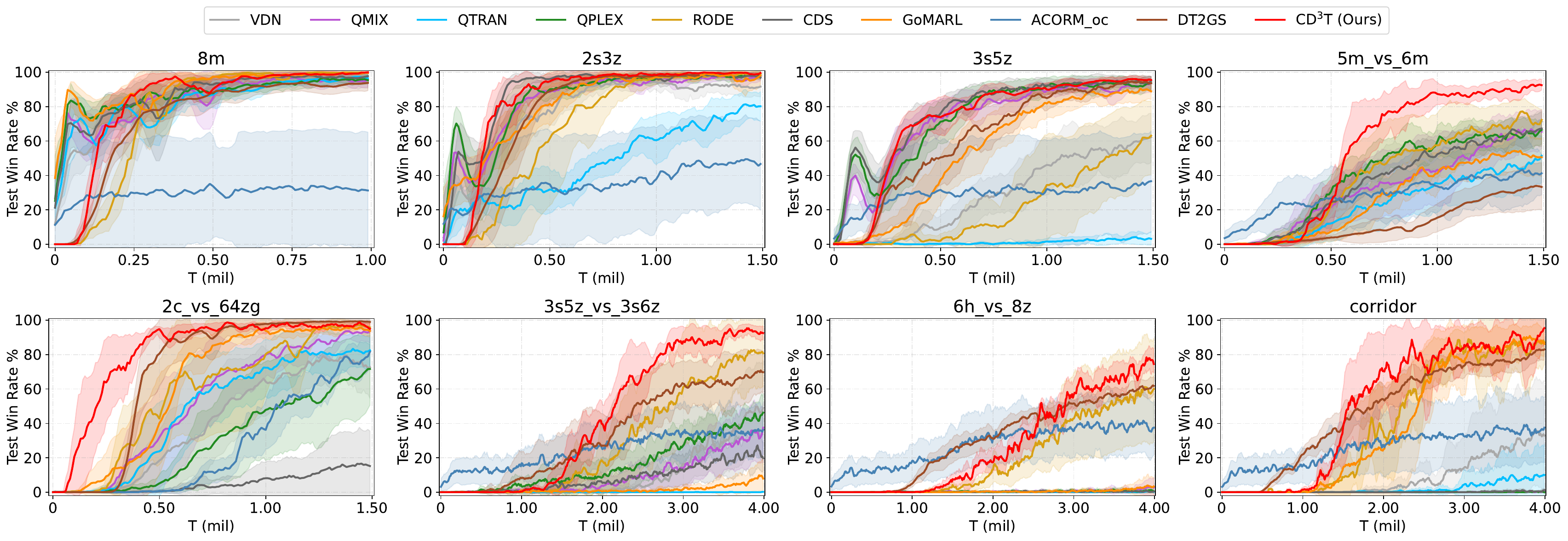}
\end{center}
\vspace{-0.3cm}
\caption{Performance comparison with baselines on \textit{easy}, \textit{hard}, and \textit{super hard} scenarios.}
\vspace{-0.3cm}
\label{fig:smacv1}
\end{figure*}

\section{Experiments}\label{experiments}
We evaluate C$\text{D}^\text{3}$T on three challenging benchmarks, including LBF~\cite{NEURIPS2020_7967cc8e}, SMAC~\cite{samvelyan2019starcraft} and SMACv2~\cite{ellis2023smacv2}.
The baselines we select are five classic value-decomposition methods (VDN~\cite{vdn}, QMIX~\cite{rashid2020weighted}, QTRAN~\cite{qtran}, QPLEX~\cite{qplex}, and CDS~\cite{li2021celebrating}), and four subtask-related methods (RODE~\cite{wang2021rode}, GoMARL~\cite{zang2023automatic}, ACORM~\cite{hu2024attention} and DT2GS~\cite{tian2023decompose}).
ACORM performs $k$-means clustering at each timestep to obtain commendable results, which undoubtedly imposes a substantial computational burden. Therefore, to improve efficiency and maintain consistency with our algorithm, we likewise apply clustering for ACORM only once at $50K$ timesteps.
We use ``ACORM$\_$oc'' to represent ACORM with once clustering in our experiments.
The implementation details of all algorithms are provided in Appendix~\ref{appendix:experimental details}, along with benchmarks and discussions on training time and scalability.
All learning curves report the mean $\pm$ standard deviation over five random seeds.

\begin{figure*}[ht]
\centering
	\subfigure[Ablation studies regarding each component of C$\text{D}^\text{3}$T.]{
		\includegraphics[width=\textwidth]{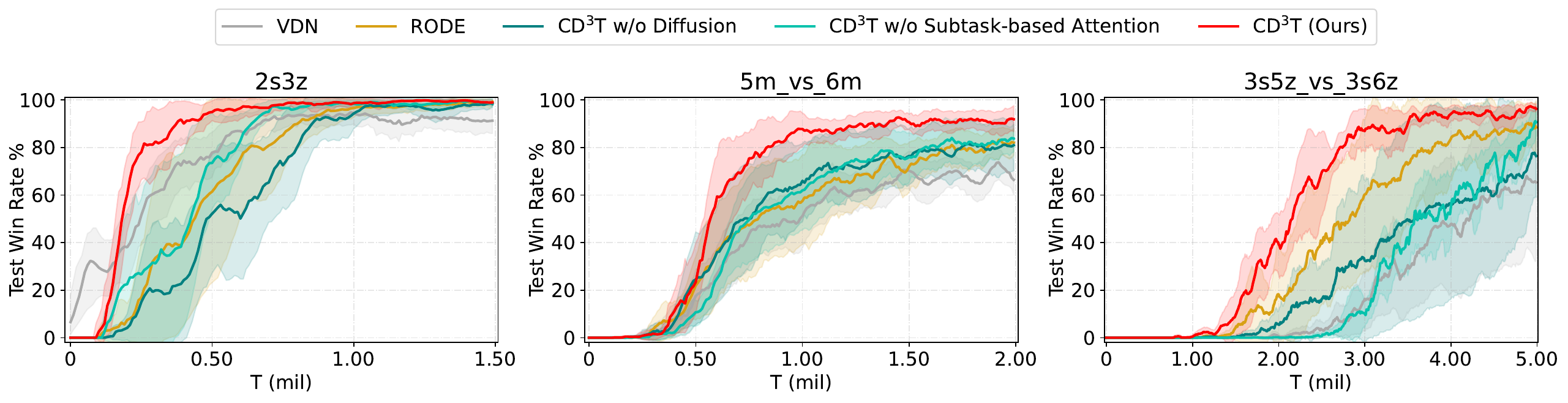}
        \label{fig:ab1}}
	\subfigure[The performance of C$\text{D}^\text{3}$T with different number of subtask clusters.]{
		\includegraphics[width=\textwidth]{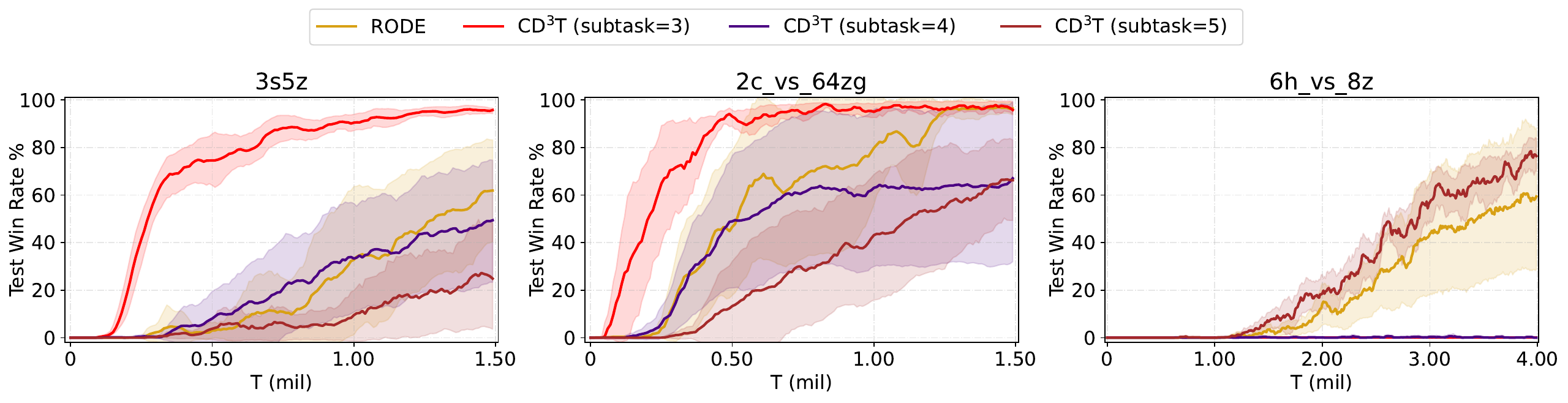}
        \label{fig:ab2}}
    \vspace{-0.4cm}
    \caption{Ablation studies of C$\text{D}^\text{3}$T on SMAC benchmark.}
    \vspace{-0.4cm}
    \label{fig:ablation}
\end{figure*}

\subsection{Performance on LBF}

We first conduct experiments on two constructed LBF tasks to assess the performance of different algorithms under two distinct settings.
\textit{Fig.~\ref{fig:lbf}} illustrates a comprehensive comparison of performance against baselines on two specially crafted LBF tasks.
Our approach shows competitive performance across LBF tasks, demonstrating its flexibility and effectiveness on both scenarios.
The failure of CDS and RODE could originate from the inability of its heterogeneous agents to effectively explore and develop collaborative strategies.
In contrast, VDN, QMIX, and QTRAN require more steps to uncover more refined policies, suggesting that they may struggle with the inherent limitations in solving spurious relationships between credit assignments and decomposed Q-values.
QPLEX receives a lower return compared to C$\text{D}^\text{3}$T between 0.1M and 0.6M timesteps, potentially necessitating additional time for exploration due to its complex value decomposition design.
C$\text{D}^\text{3}$T achieves slightly higher performance than GoMARL and ACORM, which implies that in terms of semantic representation, the action representation may have superiority over both group-based information and role representation. The marginally inferior performance of DT2GS compared to CD$^3$T further substantiates that enhancing the generalization of subtasks inevitably entails a trade-off with performance on single tasks.

\subsection{Performance on SMAC}
To further evaluate C$\text{D}^\text{3}$T, we benchmark it on the more challenging SMAC benchmark, which is a testbed commonly used for MARL algorithms.
We compare C$\text{D}^\text{3}$T with other baselines on 8 different scenarios, including \textit{easy}, \textit{hard}, and \textit{super hard} scenarios.
The details of these scenarios can be found in Appendix~\ref{appendix:benchmark}.

The experimental results for different scenarios are shown in \textit{Fig.~\ref{fig:smacv1}}.
As we can see, C$\text{D}^\text{3}$T yields almost the highest win rate on all scenarios, especially on the \textit{super hard} tasks.
QTRAN performs poorly on almost all scenarios due to its soft constraints involving two $\ell_2$-penalty terms.
Although QPLEX behaves well on easy scenarios, resulting in its tendency to fall into local optima, its performance decreases on hard scenarios.
Both VDN and QMIX can achieve satisfactory performance on some \textit{easy} or \textit{hard} maps, i.e., \texttt{8m}, \texttt{2s3z}, and \texttt{5m\_vs\_6m}, but they fail to cope with the tasks well on \textit{super hard} maps.
It should result from the fact that the super-hard task needs more efficient exploration to learn cooperation skills.
RODE fails to learn efficient policies for subtasks, which implies that its reliance solely on the simple MLP structure hinders the accurate learning of role semantics.
CDS fails to learn efficient policies since it may require more steps to explore, which celebrates diversity among agents, especially on the map \texttt{corridor} and \texttt{6h\_vs\_8z}.
GoMARL attains comparable performance with C$\text{D}^\text{3}$T on part of easy and hard maps but underperforms on two \textit{super hard} maps, possibly owing that the automatic grouping method primarily places excessive emphasis on the relative contribution of each agent to the entire group while ignoring the mutual contributions among agents within the group.
ACORM achieves strong early learning, yet once per-step clustering is removed (ACORM-oc), performance collapses on nearly all maps, implying that its contrastive role representations rely heavily on continuous reclustering.
One possible explanation for the consistently suboptimal performance of DT2GS is its excessive emphasis on generalization across a limited set of tasks.
Overall, our approach achieves impressive performance across all scenarios, which validates the advantages of C$\text{D}^\text{3}$T with its attentive design.
Additional experiments on SMACv2 in the Appendix~\ref{exp_smacv2} further confirm the effectiveness of C$\text{D}^\text{3}$T.

\subsection{Ablation Studies}
To quantify the contribution of each component, we perform three ablations and address the following questions: (a) How does the diffusion model enhance subtask representation and improve overall performance? (b) What role does the attention mechanism leveraging subtask representation play in credit assignment? (c) Does the number of subtasks affect the capability of the model?
To test components (a), we replace the diffusion model with a vanilla MLP structure and denote it as \textit{C$\text{D}^\text{3}$T w/o diffusion}. 
For (b), we replace the subtask-based attention mechanism in our mixing network, which is substituted with the QMIX method, and denote it as \textit{C$\text{D}^\text{3}$T w/o Subtask-based Attention}. 
To test component (c), we test it with the number of subtasks formed during clustering, named \textit{C$\text{D}^\text{3}$T (subtask=3)}.
Considering an excessive number of subtasks could not affect the performance for the finite action space, we set the number of subtasks to $3\leq$\textit{subtasks}$\leq5$.

The results on three scenarios with different difficulties are shown in \textit{Fig.~\ref{fig:ablation}}.
\textit{C$\text{D}^\text{3}$T w/o diffusion} attains the lowest win rates on all maps—particularly the hard and super hard ones—highlighting the critical role of diffusion-based subtask representations in high-dimensional state–action spaces.
Especially on the hard and super hard maps, it becomes clear that C$\text{D}^\text{3}$T achieves a larger margin than replacing the diffusion with a simple MLP.
The reason is that associating subtask policy with its proven powerful representational capacity in such expansive spaces benefits the performance in complex tasks.
\textit{C$\text{D}^\text{3}$T w/o Subtask-based Attention} is lower than that of C$\text{D}^\text{3}$T, which indicates the importance of subtask representation for estimating credits.
As shown in \textit{Fig.~\ref{fig:ab2}}, the performance of C$\text{D}^\text{3}$T consistently improves as the number of subtasks increases.
Generally, moderate order terms (e.g., $subtask\leq 5$) are enough for an appropriate trade-off between performance improvement and computation. 
In summary, the superior performance of C$\text{D}^\text{3}$T is conditioned on all parts, where it is largely due to the efficient subtask assignment representation.

\begin{figure*}[h]
\centering
	  \subfigure[]{
		\includegraphics[width=0.31\linewidth]{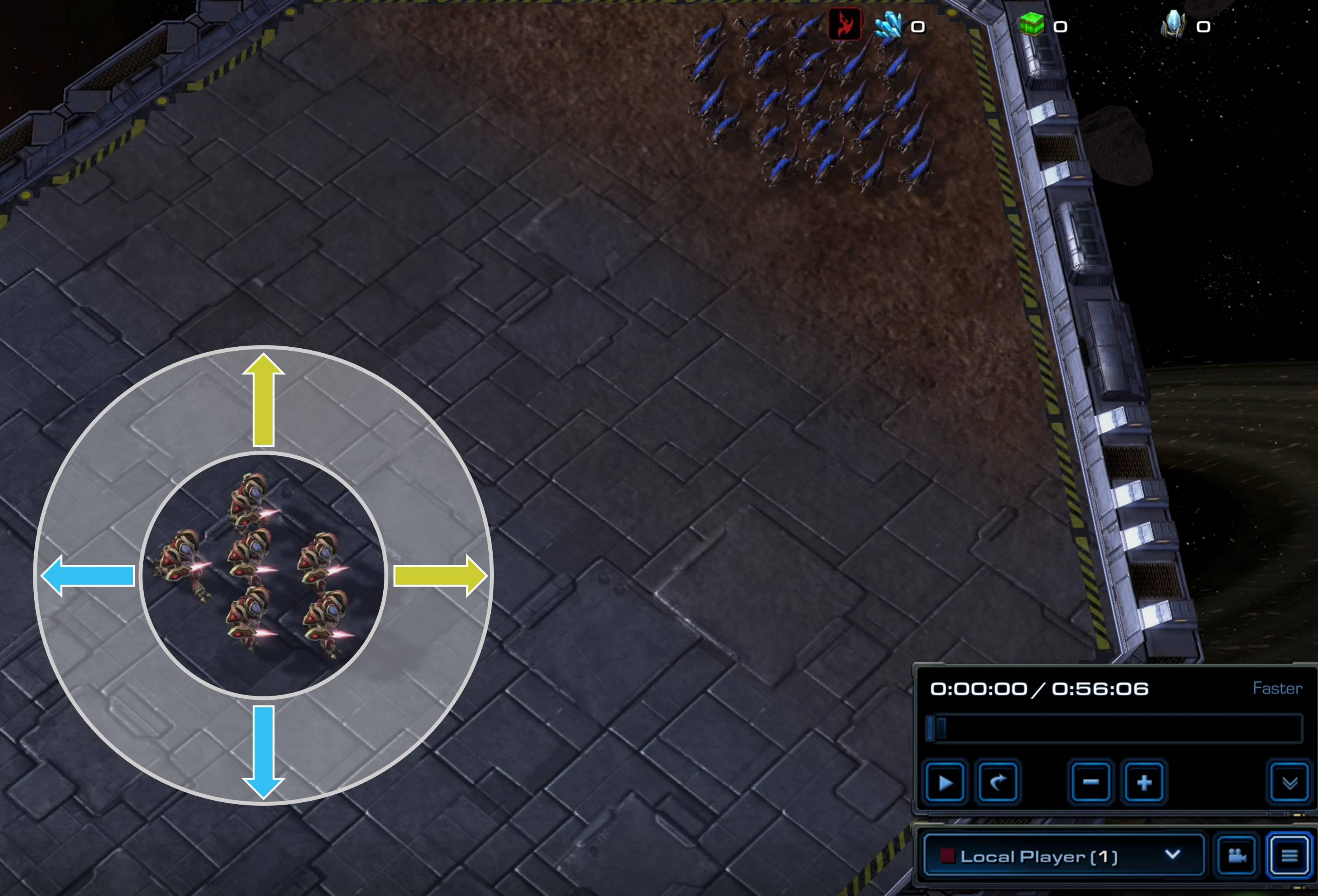}
        \label{fig:rep1}}
	  \subfigure[]{
		\includegraphics[width=0.31\linewidth]{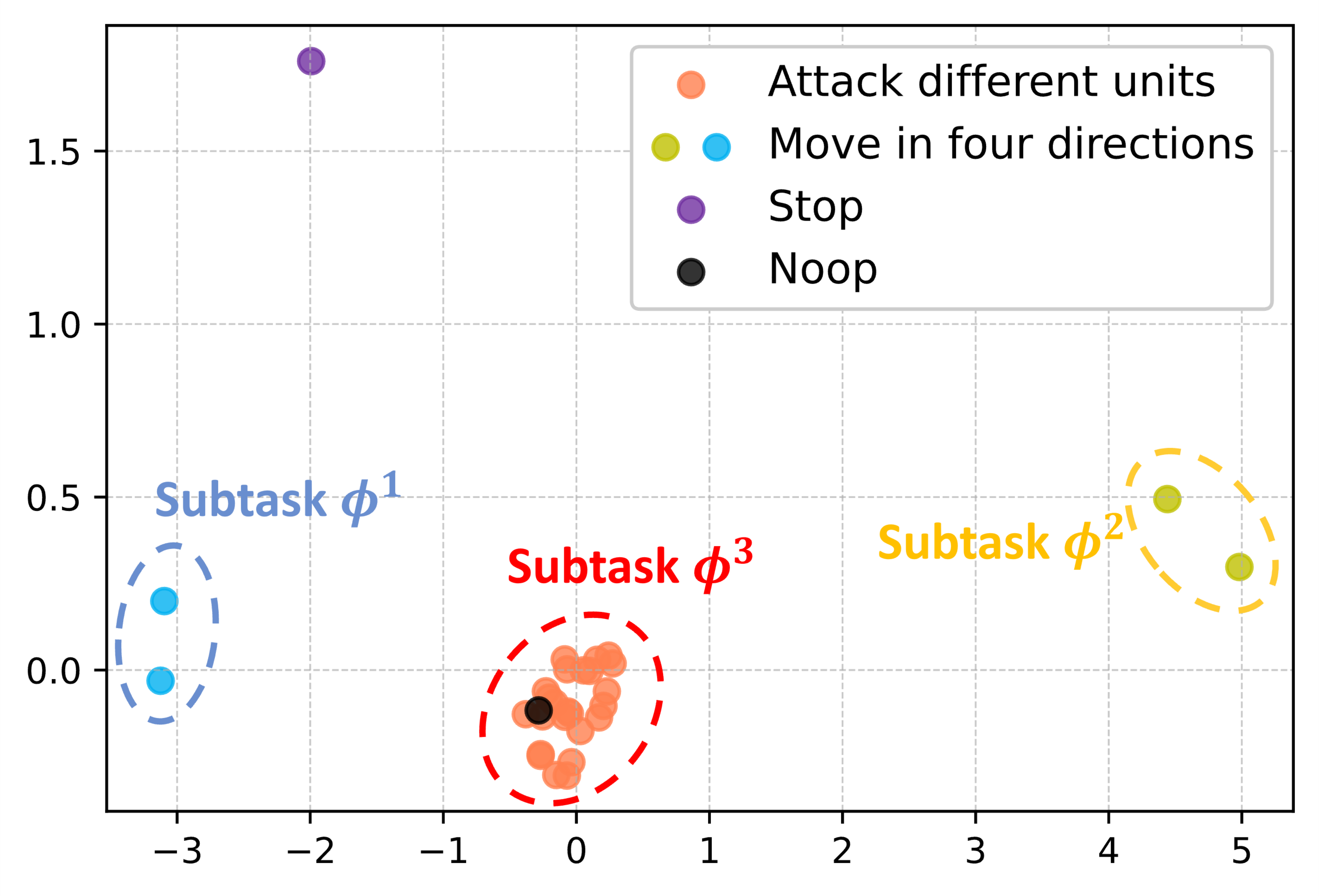}
        \label{fig:rep2}}
    	\subfigure[]{
		\includegraphics[width=0.31\linewidth]{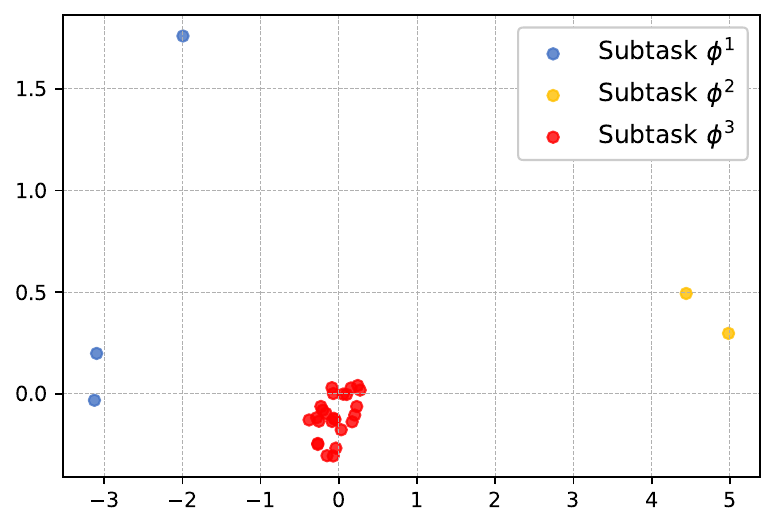}
        \label{fig:rep3}}
        \vspace{-0.28cm}
    \caption{The process of subtask generation through subtask representations. (a) illustrates that moving northward or eastward has a comparable effect on the environment by directing agents toward the enemies, whereas moving southward or westward moves agents away. (b) depicts the distribution of action representations in a two-dimensional space after PCA projection. (c) visualizes the formation of subtasks derived from action representations following clustering.}
    \vspace{-0.28cm}
    \label{fig:rep}
\end{figure*}

\begin{figure*}[ht]
\centering
	  \subfigure[Timestep = 7]{
		\includegraphics[width=0.32\linewidth]{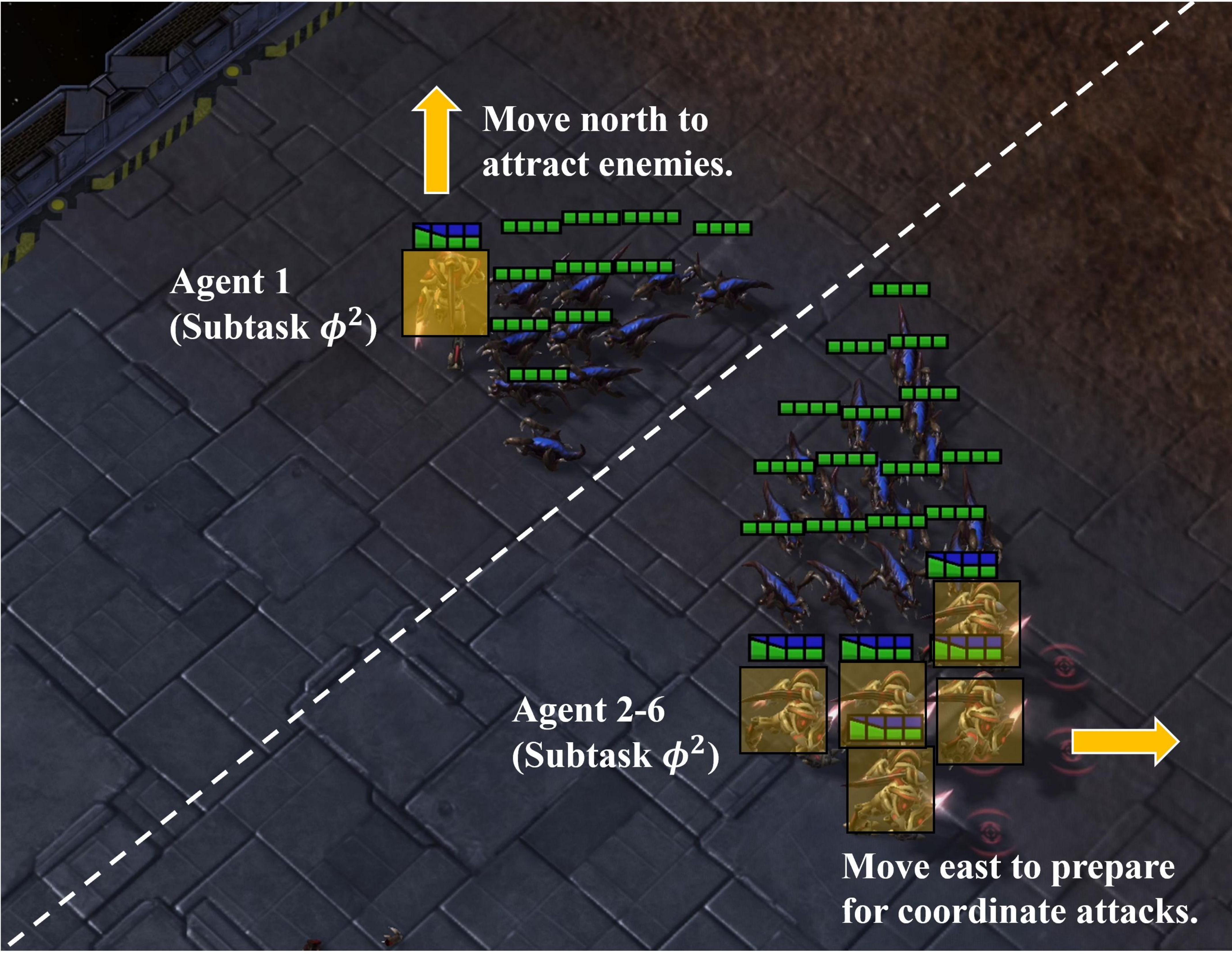}\label{fig:vis1}}
    	\subfigure[Timestep = 22]{
		\includegraphics[width=0.32\linewidth]{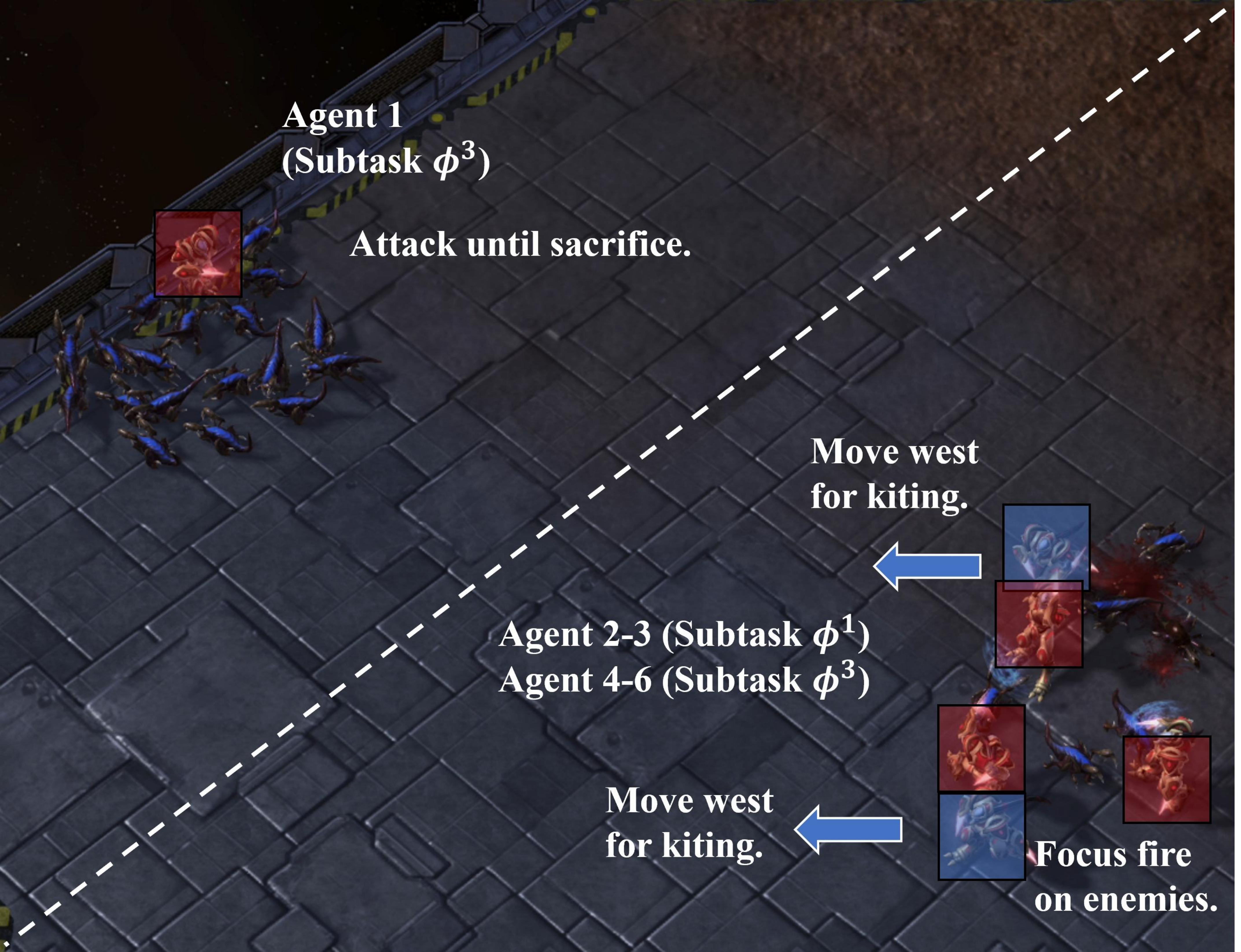}\label{fig:vis2}}
        \subfigure[Timestep = 149]{
		\includegraphics[width=0.32\linewidth]{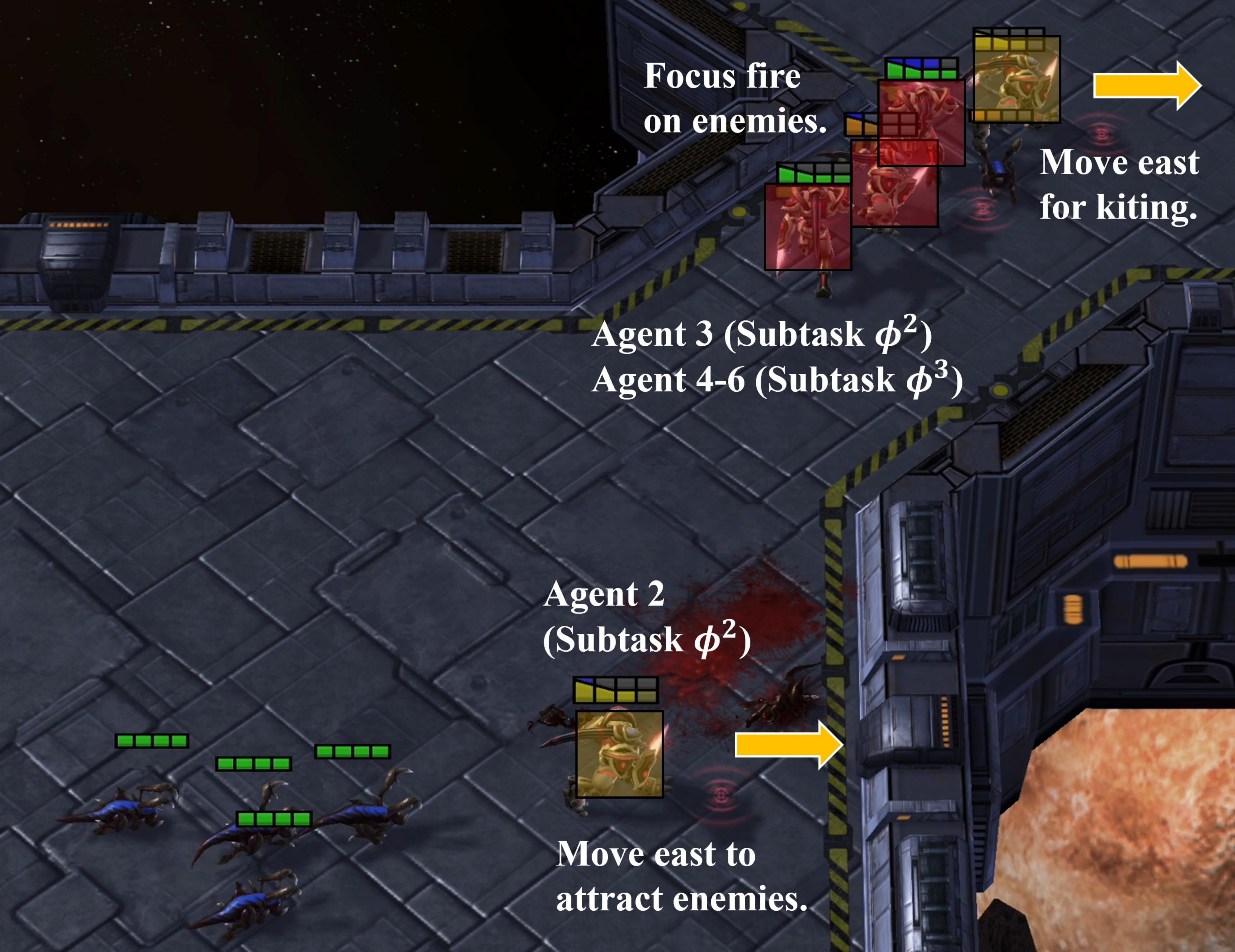}\label{fig:vis3}}
    \vspace{-0.28cm}
    \caption{Visualizations of dynamic subtask selection in one episode (351 timesteps) on \texttt{corridor}. Blue denotes subtask $\phi^1$, yellow denotes subtask $\phi^2$, and red denotes subtask $\phi^3$. (a), (b), and (c) depict game screenshots at $t = 7$, $t = 22$, and $t = 149$, respectively.}
    \vspace{-0.3cm}
    \label{fig:vis}
\end{figure*}

\subsection{Generation of Subtask Representation}

To learn action representations, we collect samples and train the diffusion model for $50K$ timesteps, guided by the loss function specified in Eq.~\eqref{eq:loss of prediction}.
The model is trained after each episode using a batch of $32$ episodes.
In \textit{Fig.~\ref{fig:rep}}, we provide an in-depth illustration of how the action representations derived from the diffusion model are strategically harnessed to enable subtask decomposition.

The \texttt{corridor} scenario features homogeneous agents and enemies—specifically, 6 Zealots versus 24 Zerglings—where all attack actions produce similar effects due to enemy uniformity. Owing to the scenario's spatial symmetry, moving north or east similarly advances agents toward the enemies, while moving south or west leads them away. The trained diffusion model captures these structural regularities within the action space, as illustrated in \textit{Fig.\ref{fig:rep1}}. In \textit{Fig.\ref{fig:rep2}}, we apply PCA to project the high-dimensional action representations into a two-dimensional space, revealing clear clusters aligned with the primary action types. These clusters emerge consistently across random seeds, demonstrating that our attentive diffusion model reliably learns subtask representations that reflect the underlying action effects.

\subsection{Visualization of Dynamic Subtask Selections}\label{sec:vis}

To gain deeper insights into the subtask selection behavior of C$\text{D}^\text{3}$T, we visualize the agent-wise subtask assignments across a representative episode (Fig.\ref{fig:vis}) and report the corresponding selection frequencies over time (Fig.\ref{fig:vis4}). 
Additional examples are included in Appendix~\ref{appendix:vis}.

A key observation is that direct engagement under numerical disadvantage is strategically suboptimal.
As shown in \textit{Fig.~\ref{fig:vis1}}, C$\text{D}^\text{3}$T assigns subtask $\phi^2$ to Agent~1 early in the episode ($t = 7$), prompting it to move northward and draw nearly half of the enemies away. This diversion enables the remaining five agents to reposition eastward for a more coordinated attack.
In \textit{Fig.~\ref{fig:vis2}}, Agent~1 switches to subtask $\phi^3$ to engage the enemies directly until eliminated. Meanwhile, Agents~2 and 3 execute kiting behaviors under $\phi^1$, while Agents~4–6 focus fire under $\phi^3$. As the battle progresses, surviving enemies regroup in the bottom-left corner, out of sight of the Zealots. During mid-phase, C$\text{D}^\text{3}$T reassigns subtask $\phi^2$ to Agent~2 to lure a subset of enemies away from the group and disrupt their formation.
Meanwhile, the remaining agents continue alternating between evasive movement and focused fire to isolate and eliminate targets.

\begin{figure}[tb]
  \centering
  \vspace{-0.3cm}
  \includegraphics[width=\linewidth]{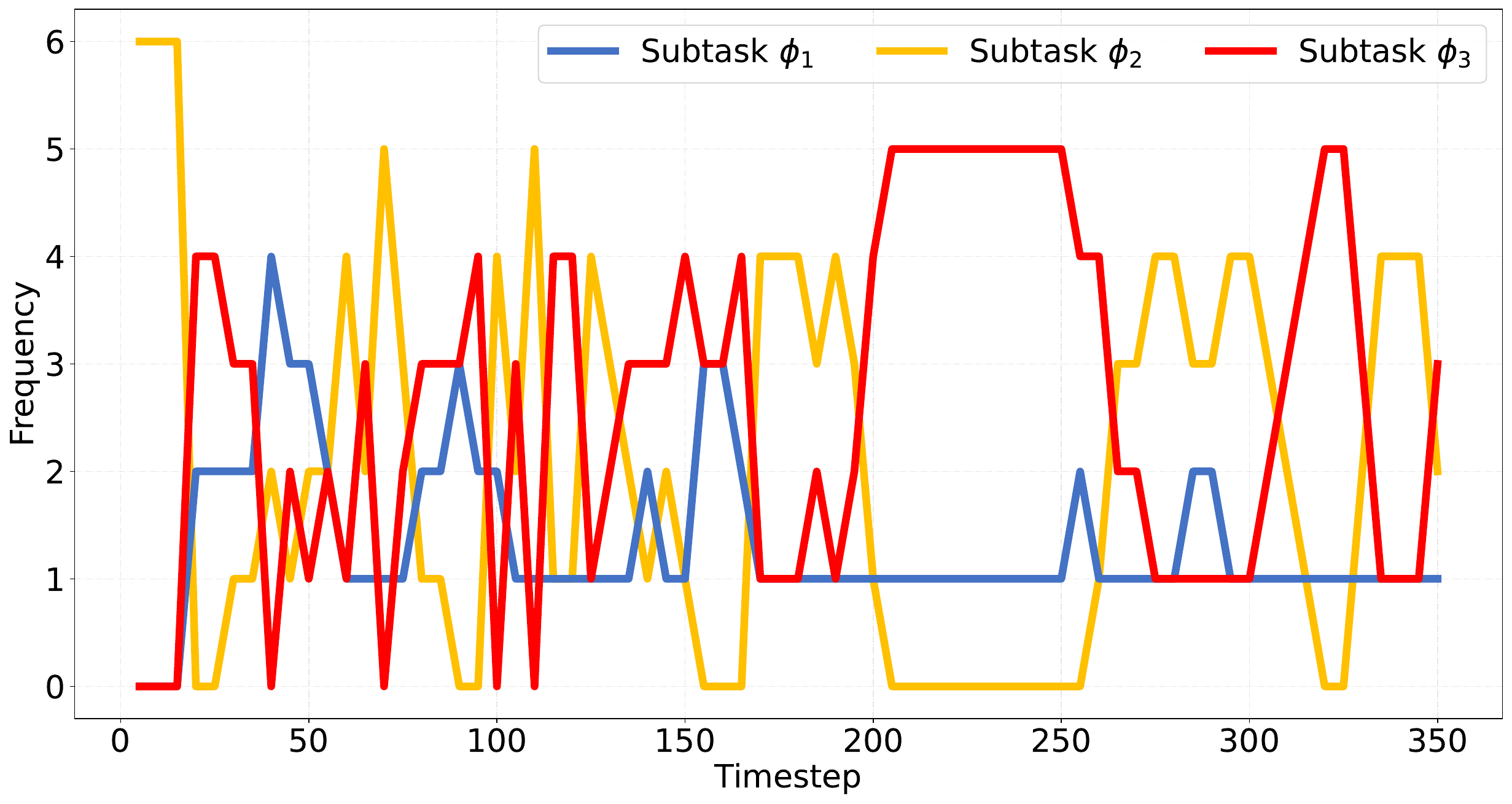}
  \caption{Demonstrates subtask selection frequencies of C$\text{D}^\text{3}$T on \texttt{corridor} map in the same episode. The curve represents the number of agents assigned the corresponding subtask at the timestep.}
  \label{fig:vis4}
  \vspace{-0.4cm}
\end{figure}

\begin{figure*}[tb]
\begin{center}
    \includegraphics[width=0.87\linewidth]{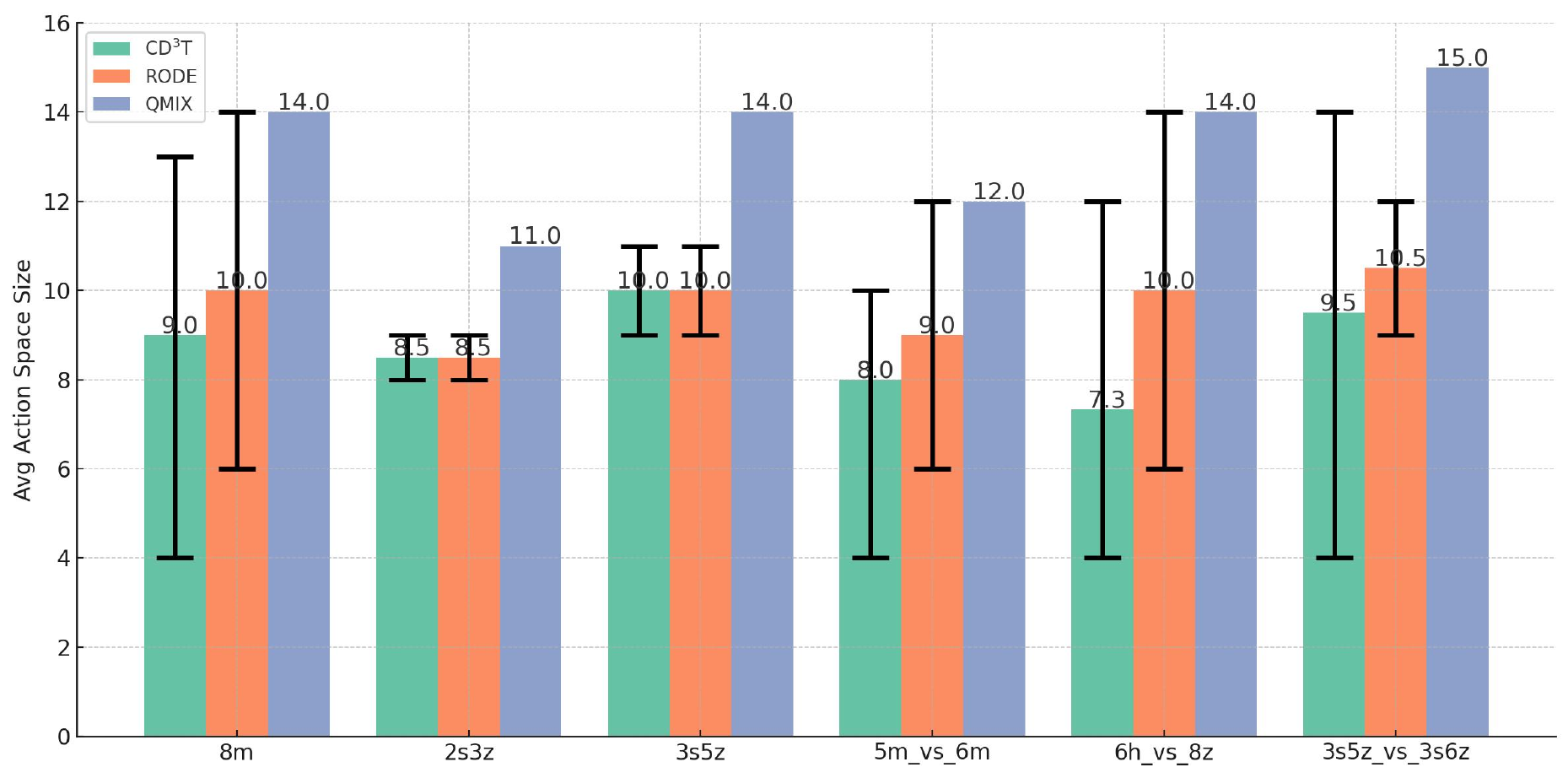}
\end{center}
\vspace{-0.3cm}
\caption{Average action space sizes of CD$^3$T, RODE, and QMIX across six SMAC scenarios. Each bar denotes the mean number of available actions per method, while the error bars capture variation across roles or agents, ranging from the minimum to the maximum number of available actions. QMIX takes actions over the full action space without any form of masking, and thus does not produce variation.
}
\vspace{-0.3cm}
\label{fig:action_space}
\end{figure*}

The frequency distribution in Fig.~\ref{fig:vis4} further confirms these dynamics.
Subtask $\phi^2$ is predominantly utilized in the early stage for diversion, followed by adaptive switching among $\phi^1$, $\phi^2$, and $\phi^3$ to sustain a balanced coordination strategy involving distraction, kiting, and concentrated attack.
This cyclical subtask adaptation underlines C$\text{D}^\text{3}$T's ability to orchestrate complex multi-agent behaviors, ultimately leading to task success.

\subsection{Empirical Visualization of Action Space Reduction}

\textit{Fig.~\ref{fig:action_space}} presents a comparative visualization of the effective action space dimensions under three methods: C$\text{D}^3$T, RODE, and QMIX.
C$\text{D}^3$T consistently achieves a more compact action space across diverse scenarios, as reflected in its lower average dimensionality and tighter confidence intervals. 
This suggests that its fine-grained subtask decomposition enables each sub-policy to operate within a task-relevant, reduced action set.
By comparison, QMIX does not incorporate any subtask or role abstraction, and thus always utilizes the full action space. 
Its action dimensionality remains fixed across scenarios, which may limit its adaptability to varying task complexity or coordination requirements.
In relatively simple environments such as \texttt{8m}, \texttt{2s3z}, and \texttt{3s5z}, both C$\text{D}^3$T and RODE exhibit similar action space sizes, indicating that in low-complexity settings, the benefits of explicit decomposition may be less pronounced.
However, in more challenging scenarios, \texttt{5m\_vs\_6m}, \texttt{6h\_vs\_8z}, and \texttt{3s5z\_vs\_3s6z}, C$\text{D}^3$T achieves more substantial reduction.
This improvement may be attributed to its explicit subtask masking mechanism, which restricts each role to a compact, specialized action subset. 
In contrast, RODE's role abstraction is more implicit and does not enforce strict action sparsity, potentially resulting in less targeted compression.

These findings highlight C$\text{D}^3$T's ability to adaptively reduce the decision space in accordance with task complexity.
By focusing each agent on a smaller, semantically meaningful set of actions, C$\text{D}^3$T not only improves computational efficiency but also facilitates more structured and scalable coordination in challenging multi-agent environments.

\section{Conclusions}\label{conclusion}
Task decomposition is a pivotal approach to simplifying complex multi-agent tasks, yet it remains a long-standing and unresolved challenge.
To address this, we proposed leveraging latent representations extracted by a diffusion model to decompose tasks into multiple subtasks.
This approach captures the relationship between subtasks and environmental dynamics more accurately.
Agents are assigned to corresponding subtasks through subtask selectors, ensuring better compatibility between agents and subtasks.
This compatibility enables agents to learn policies more efficiently in a shared learning framework.
Furthermore, by clustering latent representations, similar agents can share experiences, accelerating training and enhancing overall performance.
During training, the subtask-based attention mechanism in the mixing network effectively utilizes global state and semantic inference to guide the mixing of Q-values.
Experimental results across three benchmarks demonstrate that our method achieves superior performance in nearly all scenarios, advancing the state of the art in MARL.

\section{Acknowledgements}
This work was supported in part by the National Natural Science Foundation of China (Nos. 72394363, 72394364 \& 72394360), National Key Research and Development Program of China (2025YFA101690004), Major Science and Technology Project of Jiangsu Province under Grant BG2024041, and the Fundamental Research Funds for the Central Universities under Grant 011814380048, the China Postdoctoral Science Foundation under Grant Number 2025T180877, the Jiangsu Funding Program for Excellent Postdoctoral Talent 2025ZB267.

\section*{Impact Statements}\label{impact}
The goal of this work is to advance the field of Machine Learning.
There are no potential societal consequences of our work, none of which we feel must be specifically highlighted here.

\nocite{langley00}

\bibliography{aaai2026}


\newpage
\onecolumn
\appendix
\appsection{Diffusion Model for Latent Representation Learning}\label{appendix:diffusion}

\subsection{Forward Inference of Diffusion Models}
The forward process involves systematically adding noise to the original data, which serves as a fundamental part in comprehending the underlying principles of the diffusion model and preparing training samples essential for model development.
Given an original latent representation $\boldsymbol z_0 \sim q(x)$, the diffusion forward process adds Gaussian noise to it through times cumulatively to obtain $\boldsymbol z_1$,$\boldsymbol z_2$,...,$\boldsymbol z_K$, as shown in the following formula. Here, a series of hyperparameters of the Gaussian distribution variance $\{\beta_k\in(0,1)\}^K_{k=1}$ needs to be given. Since each timestep $k$ of the forward process is only related to the
timestep $k-1$, it can also be regarded as a Markov process:
\begin{equation}
    q\left(\boldsymbol z_k|\boldsymbol z_{k-1}\right)=\mathcal{N}\left(\boldsymbol z_k;\sqrt{1-\beta_k}\boldsymbol z_{k-1},\beta_kI\right),q\left(\boldsymbol z_{1: K} \mid \boldsymbol z_0\right)=\prod_{k=1}^K q\left(\boldsymbol z_k \mid \boldsymbol z_{k-1}\right).
\end{equation}
When $K\rightarrow\infty$
, $\boldsymbol z_K$
is a complete Gaussian noise, which is related to the choice of the mean coefficient
$\sqrt{1-\beta_k}$, and in practice $\beta_k$
increases as $t$ increases, that is $\beta_1\textless\beta_2\textless...\textless\beta_K$.

To enable gradient backpropagation through the entire process, the diffusion model adopts the reparameterization trick, commonly used in works such as \cite{jang2017categorical,kingma2014auto}. Directly sampling from a distribution, e.g., a Gaussian distribution, is non-differentiable. Since sampling $\boldsymbol z_k$ with Gaussian noise is pervasive in diffusion, reparameterization is required to ensure differentiability. Specifically, sampling $\boldsymbol z \sim \mathcal{N}(\mu_\theta, \sigma_\theta^2I)$ can be reformulated as
\begin{equation}
    \boldsymbol z=\mu_\theta+\sigma_\theta\odot\epsilon,\epsilon\in\mathcal{N}\left(0,I\right),
\end{equation}
where $\epsilon$ is an independent random variable guiding the randomness, making the process differentiable. $\boldsymbol z$ still satisfies the Gaussian distribution with mean $\mu_\theta$ and variance $\sigma_\theta^2$.
Here, $\mu_\theta$ and $\sigma_\theta^2$ can be inferred from a neural network with parameter $\theta$.
The entire sampling process is still gradient-differentiable, and the randomness is transferred to $\epsilon$.

Obtaining $\boldsymbol z_k$ through $\boldsymbol z_0$ and $\beta$ is significant in the subsequent inference and deduction of the diffusion model. At first, we assume $\alpha_k=1-\beta_k$, and $\bar{\alpha}_k=\prod_{i=1}^k\alpha_i$.
By expanding $\boldsymbol z_k$, we can then obtain
\begin{align}\notag
z_k &= \sqrt{\alpha_k} \boldsymbol z_{k-1} + \sqrt{1-\alpha_k} \epsilon_1 \quad \textit{where} \quad \epsilon_1, \epsilon_2, \dots \sim \mathcal{N}(0, \mathbf{I}); \\ \notag
    &= \sqrt{\alpha_k} \left( \sqrt{\alpha_{k-1}} \boldsymbol z_{k-2} + \sqrt{1-\alpha_{k-1}} \epsilon_2 \right) + \sqrt{1-\alpha_k} \epsilon_1 \\ \notag
    &= \sqrt{\alpha_k \alpha_{k-1}} \boldsymbol z_{k-2} + \left(\sqrt{\alpha_k(1-\alpha_{k-1})} \epsilon_2 + \sqrt{1-\alpha_k} \epsilon_1\right) \\ \notag
    &= \sqrt{\alpha_k \alpha_{k-1}} \boldsymbol z_{k-2} + \sqrt{1 - \alpha_k \alpha_{k-1}} \bar{\epsilon}_2 \quad \textit{where} \quad \bar{\epsilon}_2 \sim \mathcal{N}(0, \mathbf{I}); \\ \notag
    &= \dots \\
    &= \sqrt{\bar{\alpha}_k} \boldsymbol z_0 + \sqrt{1 - \bar{\alpha}_k} \bar{\boldsymbol z}_k.
    \label{eq14}
\end{align}
Since independent Gaussian distributions are additive, that is $\mathcal{N}\left(0,\sigma_1^2I\right)+\mathcal{N}\left(0,\sigma_2^2I\right)\sim\mathcal{N}\left(0,\left(\sigma_1^2I+\sigma_2^2\right)\right)$, we have
\begin{equation}
    \sqrt{\alpha_k(1-\alpha_{k-1})} \boldsymbol z_2 + \sqrt{1-\alpha_k} \boldsymbol z_1 \sim \mathcal{N}(0, \left[\alpha_k(1-\alpha_{k-1}) + (1-\alpha_k)\right] I)
= \mathcal{N}(0, (1-\alpha_k \alpha_{k-1}) I).
\end{equation}
Therefore, we can mix two Gaussian distributions to obtain a mixed Gaussian distribution with a standard deviation of $\sqrt{1-\alpha_k\alpha_{k-1}}$, and $\bar{\epsilon}_2$ in Eq.~\eqref{eq14} is still a standard Gaussian distribution.
Under the circumstance, $\boldsymbol z_k$ satisfies
\begin{equation}\label{eq20}
    q\left( \boldsymbol z_k|z_0 \right)=\mathcal{N}\left( \boldsymbol z_k;\sqrt{\bar{\alpha}_k}\boldsymbol z_0,\left( 1-\bar{\alpha}_k \right) I\right).
\end{equation}

\subsection{Reverse Inference of Diffusion Process}
To reconstruct the true sample from Gaussian noise, where \( \boldsymbol{z}_K \sim \mathcal{N}(\mathbf{0}, \mathbf{I}) \), one can reverse the forward process and sample from the conditional distribution \( q(\boldsymbol{z}_{k-1}|\boldsymbol{z}_k) \).
It is worth noting that if \( \beta_k \) is sufficiently small, \( q(\boldsymbol{z}_{k-1}|\boldsymbol{z}_k) \) will also follow a Gaussian distribution.
However, directly estimating \( q(\boldsymbol{z}_{k-1}|\boldsymbol{z}_k) \) is computationally intractable, as it requires access to the entire dataset.
To address this, we learn a model \( p_\theta \) to approximate these conditional probabilities, enabling the reverse diffusion process:

\begin{equation}\label{eq21}
p_\theta(\boldsymbol{z}_{0:K}) = p(\boldsymbol{z}_K) \prod_{k=1}^K p_\theta(\boldsymbol{z}_{k-1}|\boldsymbol{z}_k), \quad 
p_\theta(\boldsymbol{z}_{k-1}|\boldsymbol{z}_k) = \mathcal{N}(\boldsymbol{z}_{k-1}; \mu_\theta(\boldsymbol{z}_k, k), \Sigma_\theta(\boldsymbol{z}_k, k)),
\end{equation}

where \( \mu_\theta(\boldsymbol{z}_k, k) \) and \( \Sigma_\theta(\boldsymbol{z}_k, k) \) are parameterized by the learned model.

It is noteworthy that the reverse conditional probability becomes tractable when conditioned on \( \boldsymbol{z}_0 \):
\[
q(\boldsymbol{z}_{k-1} | \boldsymbol{z}_k, \boldsymbol{z}_0) = \mathcal{N}(\boldsymbol{z}_{k-1}; \tilde{\mu}(\boldsymbol{z}_k, \boldsymbol{z}_0), \tilde{\beta}_k \mathbf{I}).
\]

Using the rule of Bayes, the conditional probability can be expressed as
\begin{align*}
q(\boldsymbol{z}_{k-1} | \boldsymbol{z}_k, \boldsymbol{z}_0) 
&= q(\boldsymbol{z}_k | \boldsymbol{z}_{k-1}, \boldsymbol{z}_0) \frac{q(\boldsymbol{z}_{k-1} | \boldsymbol{z}_0)}{q(\boldsymbol{z}_k | \boldsymbol{z}_0)} \\
&\propto \exp \left( 
-\frac{1}{2} \left( 
\frac{(\boldsymbol{z}_k - \sqrt{\alpha_k} \boldsymbol{z}_{k-1})^2}{\beta_k} 
+ \frac{(\boldsymbol{z}_{k-1} - \sqrt{\bar{\alpha}_{k-1}} \boldsymbol{z}_0)^2}{1 - \bar{\alpha}_{k-1}} 
- \frac{(\boldsymbol{z}_k - \sqrt{\bar{\alpha}_k} \boldsymbol{z}_0)^2}{1 - \bar{\alpha}_k}
\right) \right) \\
&= \exp \left( 
-\frac{1}{2} \left( 
\left( \frac{\alpha_k}{\beta_k} + \frac{1}{1 - \bar{\alpha}_{k-1}} \right) \boldsymbol{z}_{k-1}^2 
- \left( \frac{2 \sqrt{\alpha_k}}{\beta_k} \boldsymbol{z}_k 
+ \frac{2 \sqrt{\bar{\alpha}_{k-1}}}{1 - \bar{\alpha}_{k-1}} \boldsymbol{z}_0 \right) \boldsymbol{z}_{k-1} 
+ C(\boldsymbol{z}_k, \boldsymbol{z}_0)
\right) \right),
\end{align*}

where \( C(\boldsymbol{z}_k, \boldsymbol{z}_0) \) is a term independent of \( \boldsymbol{z}_{k-1} \) and is omitted here.

Following the standard Gaussian density function, the mean and variance can be parameterized as follows, given the assumption that \( \alpha_k = 1 - \beta_k \) and \( \bar{\alpha}_k = \prod_{i=1}^k \alpha_i \) in Eq.~\eqref{eq14}:

\begin{align}\label{eq22}\notag
\tilde{\mu}_k(\boldsymbol{z}_k, \boldsymbol{z}_0) 
&= \left( \frac{\sqrt{\alpha_k}}{\beta_k} \boldsymbol{z}_k + \frac{\sqrt{\bar{\alpha}_{k-1}}}{1 - \bar{\alpha}_{k-1}} \boldsymbol{z}_0 \right) 
\Big/ \left( \frac{\alpha_k}{\beta_k} + \frac{1}{1 - \bar{\alpha}_{k-1}} \right) \\ \notag
&= \left( \frac{\sqrt{\alpha_k}}{\beta_k} \boldsymbol{z}_k + \frac{\sqrt{\bar{\alpha}_{k-1}}}{1 - \bar{\alpha}_{k-1}} \boldsymbol{z}_0 \right) 
\cdot \frac{1 - \bar{\alpha}_{k-1}}{1 - \bar{\alpha}_k} \cdot \beta_k \\ 
&= \frac{\sqrt{\alpha_k} (1 - \bar{\alpha}_{k-1})}{1 - \bar{\alpha}_k} \boldsymbol{z}_k 
+ \frac{\sqrt{\bar{\alpha}_{k-1}} \beta_k}{1 - \bar{\alpha}_k} \boldsymbol{z}_0,
\end{align}

\begin{align}\notag
\tilde{\beta}_k 
&= 1 \Big/ \left( \frac{\alpha_k}{\beta_k} + \frac{1}{1 - \bar{\alpha}_{k-1}} \right) \\ \notag
&= 1 \Big/ \left( \frac{\alpha_k - \bar{\alpha}_{k-1} + \beta_k}{\beta_k (1 - \bar{\alpha}_{k-1})} \right) \\ \notag
&= \frac{1 - \bar{\alpha}_{k-1}}{1 - \bar{\alpha}_k} \cdot \beta_k.
\end{align}

Combining the property of Eq.~\eqref{eq20}, we can represent \( \boldsymbol{z}_0 = \frac{1}{\sqrt{\bar{\alpha}_k}} (\boldsymbol{z}_k - \sqrt{1 - \bar{\alpha}_k} \boldsymbol{\epsilon}_k) \) and plug it into Eq.~\eqref{eq22}:
\begin{align}\notag
\tilde{\mu}_k 
&= 
\frac{\sqrt{\alpha_k} (1 - \bar{\alpha}_{k-1})}{1 - \bar{\alpha}_k} \boldsymbol{z}_k 
+ \frac{\sqrt{\bar{\alpha}_{k-1}} \beta_k}{1 - \bar{\alpha}_k} 
\cdot \frac{1}{\sqrt{\bar{\alpha}_k}} 
\left( \boldsymbol{z}_k - \sqrt{1 - \bar{\alpha}_k} \boldsymbol{\epsilon}_k \right) \\
&= 
\frac{1}{\sqrt{\alpha_k}} 
\left( \boldsymbol{z}_k - \frac{1 - \alpha_k}{\sqrt{1 - \bar{\alpha}_k}} \boldsymbol{\epsilon}_k \right).
\end{align}

Based on Eq.~\eqref{eq20} and Eq.~\eqref{eq21}, the proposed setup is closely aligned with the Variational Auto-encoder (VAE)~\cite{kingma2014auto} framework, allowing us to employ the variational lower bound (VLB) to optimize the negative log-likelihood:
\begin{align*}
-\log p_\theta(\boldsymbol{z}_0) 
&\leq -\log p(\boldsymbol{z}_0) + D_\text{KL}(q(\boldsymbol{z}_{1:K}|\boldsymbol{z}_0) \| p_\theta(\boldsymbol{z}_{1:K}|\boldsymbol{z}_0)) \\
&= -\log p(\boldsymbol{z}_0) + \mathbb{E}_{\boldsymbol{z}_{1:K} \sim q(\boldsymbol{z}_{1:K}|\boldsymbol{z}_0)} 
\left[ \log \frac{q(\boldsymbol{z}_{1:K}|\boldsymbol{z}_0)}{p_\theta(\boldsymbol{z}_{1:K}|\boldsymbol{z}_0)/p_\theta(\boldsymbol{z}_0)} \right] \\
&= -\log p(\boldsymbol{z}_0) + \mathbb{E}_q 
\left[ \log \frac{q(\boldsymbol{z}_{1:K}|\boldsymbol{z}_0)}{p_\theta(\boldsymbol{z}_{1:K}|\boldsymbol{z}_0)} \right] 
+ \log p_\theta(\boldsymbol{z}_0) \\
&= \mathbb{E}_q \left[ \log \frac{q(\boldsymbol{z}_{1:K}|\boldsymbol{z}_0)}{p_\theta(\boldsymbol{z}_0:\boldsymbol{z}_K)} \right].
\end{align*}

\[
\textit{Let } \mathcal L_\textit{VLB} = \mathbb{E}_{q(\boldsymbol{z}_{0:K})} 
\left[ \log \frac{q(\boldsymbol{z}_{1:K}|\boldsymbol{z}_0)}{p_\theta(\boldsymbol{z}_{0:K})} \right] 
\geq -\mathbb{E}_{q(\boldsymbol{z}_0)} \log p_\theta(\boldsymbol{z}_0).
\]

We can also derive the same result by applying Jensen's inequality. Suppose our goal is to minimize the cross-entropy as the learning objective:
\begin{align*}
\mathcal L_\textit{CE} 
&= -\mathbb{E}_{q(\boldsymbol{z}_0)} \log p_\theta(\boldsymbol{z}_0) \\
&= -\mathbb{E}_{q(\boldsymbol{z}_0)} \log \left( \int p_\theta(\boldsymbol{z}_{0:K}) \, d\boldsymbol{z}_{1:K} \right) \\
&= -\mathbb{E}_{q(\boldsymbol{z}_0)} \log \left( \int q(\boldsymbol{z}_{1:K}|\boldsymbol{z}_0) 
\frac{p_\theta(\boldsymbol{z}_{0:K})}{q(\boldsymbol{z}_{1:K}|\boldsymbol{z}_0)} \, d\boldsymbol{z}_{1:K} \right) \\
&= -\mathbb{E}_{q(\boldsymbol{z}_0)} \log \left( \mathbb{E}_{q(\boldsymbol{z}_{1:K}|\boldsymbol{z}_0)} 
\left[ \frac{p_\theta(\boldsymbol{z}_{0:K})}{q(\boldsymbol{z}_{1:K}|\boldsymbol{z}_0)} \right] \right) \\
&\leq -\mathbb{E}_{q(\boldsymbol{z}_{0:K})} \log \left( 
\frac{p_\theta(\boldsymbol{z}_{0:K})}{q(\boldsymbol{z}_{1:K}|\boldsymbol{z}_0)} \right) \\
&= \mathbb{E}_{q(\boldsymbol{z}_{0:K})} 
\left[ \log \frac{q(\boldsymbol{z}_{1:K}|\boldsymbol{z}_0)}{p_\theta(\boldsymbol{z}_{0:K})} \right] 
= \mathcal L_\textit{VLB}.
\end{align*}

To convert each term in the equation to be analytically computable, the objective can be further rewritten to be a combination of several KL-divergence and entropy terms~\cite{sohldickstein2015deepunsupervisedlearningusing}:
\begin{align}\label{eq24}
\mathcal{L}_\textit{VLB} \notag
&= \mathbb{E}_{q(\boldsymbol{z}_{0:K})} \left[ \log \frac{q(\boldsymbol{z}_{1:K}|\boldsymbol{z}_0)}{p_\theta(\boldsymbol{z}_{0:K})} \right] \\ \notag
&= \mathbb{E}_q \left[ \log \frac{\prod_{k=1}^K q(\boldsymbol{z}_k|\boldsymbol{z}_{k-1})}{p_\theta(\boldsymbol{z}_K) \prod_{k=1}^K p_\theta(\boldsymbol{z}_{k-1}|\boldsymbol{z}_k)} \right] \\ \notag
&= \mathbb{E}_q \left[ -\log p_\theta(\boldsymbol{z}_K) + \sum_{k=1}^K \log \frac{q(\boldsymbol{z}_k|\boldsymbol{z}_{k-1})}{p_\theta(\boldsymbol{z}_{k-1}|\boldsymbol{z}_k)} \right] \\ \notag
&= \mathbb{E}_q \left[ -\log p_\theta(\boldsymbol{z}_K) + \sum_{k=2}^K \log \frac{q(\boldsymbol{z}_k|\boldsymbol{z}_{k-1})}{p_\theta(\boldsymbol{z}_{k-1}|\boldsymbol{z}_k)} + \log \frac{q(\boldsymbol{z}_1|\boldsymbol{z}_0)}{p_\theta(\boldsymbol{z}_0|\boldsymbol{z}_1)} \right] \\ \notag
&= \mathbb{E}_q \left[ -\log p_\theta(\boldsymbol{z}_K) + \sum_{k=2}^K \log \frac{q(\boldsymbol{z}_k|\boldsymbol{z}_{k-1}, \boldsymbol{z}_0)}{p_\theta(\boldsymbol{z}_{k-1}|\boldsymbol{z}_k)} + \log \frac{q(\boldsymbol{z}_K|\boldsymbol{z}_0)}{p_\theta(\boldsymbol{z}_K)} \right] \\ \notag
&= \mathbb{E}_q \left[ \log \frac{q(\boldsymbol{z}_K|\boldsymbol{z}_0)}{p_\theta(\boldsymbol{z}_K)} + \sum_{k=2}^K \log \frac{q(\boldsymbol{z}_k|\boldsymbol{z}_{k-1}, \boldsymbol{z}_0)}{p_\theta(\boldsymbol{z}_{k-1}|\boldsymbol{z}_k)} - \log p_\theta(\boldsymbol{z}_0|\boldsymbol{z}_1) \right] \\ 
&= \mathbb{E}_q \left[ \underbrace{D_\text{KL}(q(\boldsymbol{z}_K|\boldsymbol{z}_0) \| p_\theta(\boldsymbol{z}_K))}_{\mathcal{L}_K} 
+ \sum_{k=2}^K \underbrace{D_\text{KL}(q(\boldsymbol{z}_{k-1}|\boldsymbol{z}_k, \boldsymbol{z}_0) \| p_\theta(\boldsymbol{z}_{k-1}|\boldsymbol{z}_k))}_{\mathcal{L}_{k-1}} 
\underbrace{- \log p_\theta(\boldsymbol{z}_0|\boldsymbol{z}_1)}_{\mathcal L_0} \right].
\end{align}

The variational lower bound can be expressed using the labels from Eq.~\eqref{eq24} as follows:
\begin{equation*}
    \mathcal{L}_\textit{VLB}=\mathcal L_{K}+\sum\limits_{k=1}^{K-1}\mathcal{L}_{k}+\mathcal{L}_0.
\end{equation*}

Every KL term in $\mathcal{L}_\text{VLB}$ (except for $\mathcal{L}_0$) compares two Gaussian distributions, and therefore they can be computed in a closed form. $\mathcal{L}_K$ is constant and can be ignored during training because $q$ has no learnable parameters, and $\boldsymbol{z}_K$ is Gaussian noise. Previous method~\cite{ho2020denoising} models $\mathcal{L}_0$ using a separate discrete decoder derived from 
$\mathcal{N}(\boldsymbol{z}_0; \mu_\theta(\boldsymbol{z}_1, 1), \Sigma_\theta(\boldsymbol{z}_1, 1))$.

\subsection{Parameterization of $L_k$ for Training Loss}
In order to train a neural network to approximate the conditional probability distributions from Eq.~\eqref{eq21} in the reverse diffusion process: 
\[
p_\theta(\boldsymbol{z}_{k-1} | \boldsymbol{z}_k) = \mathcal{N}(\boldsymbol{z}_{k-1}; \mu_\theta(\boldsymbol{z}_k, k), \Sigma_\theta(\boldsymbol{z}_k, k)),
\]
we adopt the approach of training \(\mu_\theta\) to predict \(\tilde{\mu}_k=\frac{1}{\sqrt{\alpha_k}}\left(\boldsymbol{z}_k-\frac{1-\alpha_k}{\sqrt{1-\bar{\alpha}_k}}\boldsymbol{\epsilon}_k\right)\). Given that \(\boldsymbol{z}_k\) is accessible as input during training, we can reparameterize the Gaussian noise term to directly predict \(\boldsymbol{\epsilon}_k\) from the input \(\boldsymbol{z}_k\) at timestep \(k\):
\begin{align*}
\mu_\theta(\boldsymbol{z}_k, k) 
&= \frac{1}{\sqrt{\alpha_k}} 
\left( \boldsymbol{z}_k - \frac{1 - \alpha_k}{\sqrt{1 - \bar{\alpha}_k}} \boldsymbol{\epsilon}_\theta(\boldsymbol{z}_k, k) \right), \\
\textit{thus}\ \boldsymbol{z}_{k-1} 
&= \mathcal{N} \left( 
\boldsymbol{z}_{k-1}; 
\frac{1}{\sqrt{\alpha_k}} 
\left( \boldsymbol{z}_k - \frac{1 - \alpha_k}{\sqrt{1 - \bar{\alpha}_k}} \boldsymbol{\epsilon}_\theta(\boldsymbol{z}_k, k) \right), 
\Sigma_\theta(\boldsymbol{z}_k, k)
\right).
\end{align*}
The loss term $\mathcal{L}_k$ is parameterized to minimize the difference from $\tilde{\mu}$:
\begin{align}\label{eq25}\notag
\mathcal{L}_k 
&= \mathbb{E}_{\boldsymbol{z}_0, \boldsymbol{\epsilon}} 
\left[ 
\frac{1}{2 \| \Sigma_\theta(\boldsymbol{z}_k, k) \|_2^2} 
\left\| \tilde{\mu}_k(\boldsymbol{z}_k, \boldsymbol{z}_0) - \mu_\theta(\boldsymbol{z}_k, k) \right\|^2 
\right] \\ \notag
&= \mathbb{E}_{\boldsymbol{z}_0, \boldsymbol{\epsilon}} 
\left[ 
\frac{1}{2 \| \Sigma_\theta \|_2^2} 
\left\| \frac{1}{\sqrt{\alpha_k}} 
\left( \boldsymbol{z}_k - \frac{1 - \alpha_k}{\sqrt{1 - \bar{\alpha}_k}} \boldsymbol{\epsilon} \right) 
- \frac{1}{\sqrt{\alpha_k}} 
\left( \boldsymbol{z}_k - \frac{1 - \alpha_k}{\sqrt{1 - \bar{\alpha}_k}} \boldsymbol{\epsilon}_\theta(\boldsymbol{z}_k, k) \right) 
\right\|^2 
\right] \\ \notag
&= \mathbb{E}_{\boldsymbol{z}_0, \boldsymbol{\epsilon}} 
\left[ 
\frac{(1 - \alpha_k)^2}{2 \alpha_k (1 - \bar{\alpha}_k) \| \Sigma_\theta \|_2^2} 
\left\| \boldsymbol{\epsilon} - \boldsymbol{\epsilon}_\theta(\boldsymbol{z}_k, k) \right\|^2 
\right] \\
&= \mathbb{E}_{\boldsymbol{z}_0, \boldsymbol{\epsilon}} 
\left[ 
\frac{(1 - \alpha_k)^2}{2 \alpha_k (1 - \bar{\alpha}_k) \| \Sigma_\theta \|_2^2} 
\left\| \boldsymbol{\epsilon} - \boldsymbol{\epsilon}_\theta 
\left( \sqrt{\bar{\alpha}_k} \boldsymbol{z}_0 + \sqrt{1 - \bar{\alpha}_k} \boldsymbol{\epsilon}, k \right) 
\right\|^2 
\right].
\end{align}
Empirically, the training procedure works better with a simplified objective from Eq.~\eqref{eq25} that ignores the weighting term:
\begin{align*}
\mathcal{L}_d(\theta) 
&= \mathbb{E}_{k \sim [1, K], \boldsymbol{z}_0, \boldsymbol{\epsilon}_k} 
\left[ \left\| \boldsymbol{\epsilon}_k - \boldsymbol{\epsilon}_\theta(\boldsymbol{z}_k, k) \right\|^2 \right] \\
&= \mathbb{E}_{k \sim [1, K], \boldsymbol{z}_0, \boldsymbol{\epsilon}_k} 
\left[ \left\| \boldsymbol{\epsilon}_k - \boldsymbol{\epsilon}_\theta 
\left( \sqrt{\bar{\alpha}_k} \boldsymbol{z}_0 + \sqrt{1 - \bar{\alpha}_k} \boldsymbol{\epsilon}_k, k \right) 
\right\|^2 \right]\\
&= \mathbb{E}_{k \sim [1, K], \boldsymbol{z}_0, \boldsymbol{\epsilon}_k} 
\left[ \left\| \boldsymbol{\epsilon}_k - \boldsymbol{\epsilon}_\theta 
\left( \sqrt{\bar{\alpha}_k} \boldsymbol{z}_0 + \sqrt{1 - \bar{\alpha}_k} \boldsymbol{\epsilon}_k, k \right) 
\right\|^2 \right].
\end{align*}

\appsection{Theoretical Analysis of Subtask-based Value Decomposition}\label{appendix:credit assignment}

In this section, we theoretically derive the expansion formula for each individual Q-value function and further explore the non-linear combination of multiple individual Q-values into a global Q-value under the multi-agent value decomposition framework.

Following the general framework of MARL, the Q-value $Q_{tot}(s, \boldsymbol{a})$ is a function on the state vector 
$s$ and joint actions $\boldsymbol{a} = (a_1, \dots, a_N)$. By applying the implicit function theorem, $Q_{tot}$ can also be 
viewed as a function in terms of individual Q-value $Q_i$:
\[
Q_{tot} = Q_{tot}(s, Q_1, Q_2, \dots, Q_N), \ \textit{where } Q_i = Q_i(s, a_i) \approx Q_i(\tau_i, a_i).
\]

To simplify our analysis, we assume that no agent operates independently of the group. In this context, an agent 
$i$ either contributes to the group or is treated as an individual optimizing its own policy. Mathematically, this ensures that $\frac{\partial Q_{tot}}{\partial Q_i}$ is not identically zero, meaning that variations in $Q_i$ generally influence $Q_{tot}$, though it may vanish at certain points:
\[
\frac{\partial Q_{tot}}{\partial Q_i} \neq 0.
\]

We investigate the local behaviors of $Q_{tot}$ and $Q_i$ nearly a maximum point $\boldsymbol{a}$ 
in the action space assuming the state $s$ is fixed. Since the gradient $\frac{\partial Q_{tot}}{\partial \boldsymbol{a}}$ vanishes at the optimum 
point $\boldsymbol{a}_o$, we have
\[
\frac{\partial Q_{tot}}{\partial a_i} = \frac{\partial Q_{tot}}{\partial Q_i} \frac{\partial Q_i}{\partial a_i} = 0.
\]
We conclude that
\[
\frac{\partial Q_i}{\partial a_i}(a^o) = 0.
\]
Consequently, we have local expansion of $Q_i(a_i)$ as follows:

\begin{equation}\label{eq26}
    Q_i(a_i) = \alpha_i + \beta_i(a_i - a^o_i)^2 + o((a_i - a^o_i)^2),
\end{equation}
where $\alpha_i$ and $\beta_i$ are constants. Thereafter, we can theoretically derive that the non-linear dependence of the global Q-value $Q_{tot}$ on individual Q-value $Q_i$ (near a maximal point $\boldsymbol{a}^o$) as shown in the following Theorem~\ref{thm1}.

\begin{theorem}\label{thm1}
Assume that the action space is continuous and there is no independent agent. Then there exist constants $c(s)$, $\lambda_{i,h}(s)$ (depending on state $s$), such that the local expansion of $Q_{tot}$ admits the following form:
\begin{equation}\label{eq27}
    Q_{tot}(s, \boldsymbol{a}) \approx c(s) + \sum_{i,h} \lambda_{i,h}(s) Q_i(\tau_i, a_i).
\end{equation}
where $\lambda_{i,h}$ is a linear functional of all partial derivatives $\frac{\partial^h Q_{tot}}{\partial Q_{i_1} \dots \partial Q_{i_h}}$ of order $h$, and decays super-exponentially fast in $h$.

\end{theorem}

\textit{Proof.} We expand $Q_{tot}$ in terms of $Q_i$:
\begin{equation}\label{eq28}
Q_{tot} = c + \sum_i \mu_i Q_i + \sum_{ij} \mu_{ij} Q_i Q_j + \dots + \sum_{i_1,\dots,i_k} \mu_{i_1,\dots,i_k} Q_{i_1} \dots Q_{i_k} + \dots
\end{equation}
where
\[
\mu_i = \frac{\partial Q_{tot}}{\partial Q_i}, \quad \mu_{ij} = \frac{1}{2} \frac{\partial^2 Q_{tot}}{\partial Q_i \partial Q_j},
\]
and in general
\[
\mu_{i_1,\dots,i_k} = \frac{1}{k!} \frac{\partial^k Q_{tot}}{\partial Q_{i_1} \dots \partial Q_{i_k}}.
\]

Recall Eq.~\eqref{eq26} for each individual Q value, we have a local expansion
\[
Q_i(a_i) = \alpha_i + \beta_i (a_i - a^o_i)^2 + o((a_i - a^o_i)^2).
\]

Now we apply the above equation to the second-order term in Eq.~\eqref{eq27}:
\[
\begin{aligned}
    \sum_{i,j} \mu_{ij} Q_i Q_j 
    &= \sum_{i,j} \mu_{ij} \left( \alpha_i + \beta_i (a_i - a_i^o)^2 \right) \left( \alpha_j + \beta_j (a_j - a_j^o)^2 \right) + o(\|a - a^o\|^2) \\
    &= \sum_{i,j} \mu_{ij} \alpha_i \alpha_j + 2 \sum_{i,j} \mu_{ij} \alpha_i \beta_j (a_j - a_j^o)^2 + o(\|a - a^o\|^2) \\
    &= \sum_{i,j} \mu_{ij} \alpha_i \alpha_j + 2 \sum_{i,j} \mu_{ij} \alpha_j (Q_i - \alpha_i) + o(\|a - a^o\|^2) \\
    &= - \sum_{i,j} \mu_{ij} \alpha_i \alpha_j + 2 \sum_{i,j} \mu_{ij} \alpha_j Q_i + o(\|a - a^o\|^2).
\end{aligned}
\]
Therefore, we will take
\[
\lambda_{i,2} = 2 \sum_j \mu_{ij} \alpha_j.
\]
In general, we have
\[
\mu_{i_1,\dots,i_k} = \frac{1}{k!} \frac{\partial^k Q_{tot}}{\partial Q_{i_1} \dots \partial Q_{i_k}},
\]
and
\[
\sum_{i_1,\dots,i_k} \mu_{i_1,\dots,i_k} Q_{i_1} \dots Q_{i_k} = (k-1) \sum_{i_1,\dots,i_k} \mu_{i_1,\dots,i_k} \alpha_{i_1} \dots \alpha_{i_k-1}.
\]
Hence, we take
\[
\lambda_{i,k} = k \sum_{i_1,\dots,i_k} \mu_{i_1,\dots,i_k} \alpha_{i_1} \dots \alpha_{i_k-1}.
\]

The convergence of the series $\sum_k \lambda_{i,k}$ only requires mild conditions, for example, boundedness or even small growth of partial derivatives $\frac{\partial^k Q_{tot}}{\partial Q^1 \dots \partial Q^k}$ in terms of $k$. Hence, Theorem~\ref {thm1} is proved.

We use multiple attention heads to implement the approximations of different
orders of partial derivatives. By summing up the head Q-values $Q_h$ from different heads, we obtain
\begin{equation}
Q_{tot} \approx c(s) + \sum_{h=1}^H Q_h, \quad \textit{where } Q_h = \sum_{i=1}^N \lambda_{i,h} Q_i,
\end{equation}
\textit{H} is the number of attention heads. Lastly, the first term \( c(s) \) in Eq.~\eqref{eq27} could be learned by a neural network with the global state \( \boldsymbol s \) as the input. Our method naturally holds the monotonicity and then achieves the IGM property between \( Q_{tot} \) and \( Q_i \):
\[
\frac{\partial Q_{tot}}{\partial Q_i} \geq 0, \forall i \in \{1, 2, \dots, N\}.
\]
Thus, C$\text{D}^\text{3}$T allows tractable maximization of the joint action-value in off-policy learning and
guarantees consistency between the centralized and decentralized policies.

\appsection{Pseudo Code}\label{app:pseudo code}
\begin{algorithm}
\caption{C$\text{D}^\text{3}$T in MARL}
\textbf{Input:} $B$: The Boolean variable whether to update the action representation\\
\hspace*{2.7em} $J$: Number of clusters\\
\hspace*{2.7em} $\Delta T$: Time interval for updating subtask selector\\
\hspace*{2.7em} $N$: Number of agents\\
\hspace*{2.7em} $\mathcal{D}$: Replay buffer\\
\hspace*{2.7em} $T$: Timesteps of a learning episode\\
\hspace*{2.7em} $T_{tot}$: Total timesteps of learning

\begin{algorithmic}[1]
\STATE Initialize all network parameters
\STATE Initialize the replay buffer $\mathcal{D}$ for storing agent trajectories
\FOR {episode = $1, 2, \dots$}
    \STATE Initialize history agent embedding $h_i^0$ and action vector $a_i^0$ for each agent
    \FOR{$t = 1, 2, \dots, T$}
        \STATE Obtain each agent's partial observation $\{o_i^t\}_{i=1}^N$ and global state $s_t$
        \IF {$B$ = True}
            \IF{$T_{tot}\textless50K$}
                \FOR{agent $i = 1, 2, \dots, N$}
                    \STATE Calculate the action representaion $z_{a_i}^t$ 
                     \STATE Update diffusion model according to Eq.~\eqref{eq:loss of prediction} 
                \ENDFOR
            \ELSE
                \STATE Partition action representations $\{z_{a_i}^t\}_{i=1}^N$ into $J$ clusters $\{\phi^j\}_{j=1}^J$ using $k$-means
                \STATE Calculate the subtask representaion $z_{\phi^j}^t$
                \STATE $B$ = False
            \ENDIF
        \ENDIF
        
    \STATE Execute joint action $a^t = [a_1^t, a_2^t, \dots, a_n^t]^\top$ and obtain global reward $r^t$
    \ENDFOR
    \STATE Store the trajectory to $\mathcal{D}$
    \STATE Sample a batch of trajectories from $\mathcal{D}$
    \IF{$T_{tot}$ mod $\Delta T == 0$}
        \STATE Select subtask $\phi_j$ for each agent
    \STATE Update the parameters of subtask selector Q-network and the mixing network according to Eq.\eqref{eq: update subtask selector}
    \ENDIF
    \STATE Update the parameters of the individual Q-network and the mixing network according to Eq.\eqref{eq: update subtask poicy}
\ENDFOR
\end{algorithmic}
\end{algorithm}

\appsection{Experimental Details}\label{appendix:experimental details}

\subsection{Benchmark and settings}\label{appendix:benchmark}

\begin{figure*}[ht]
\centering
	  \subfigure[Level Based Foraging]{
		\includegraphics[width=0.22\linewidth]{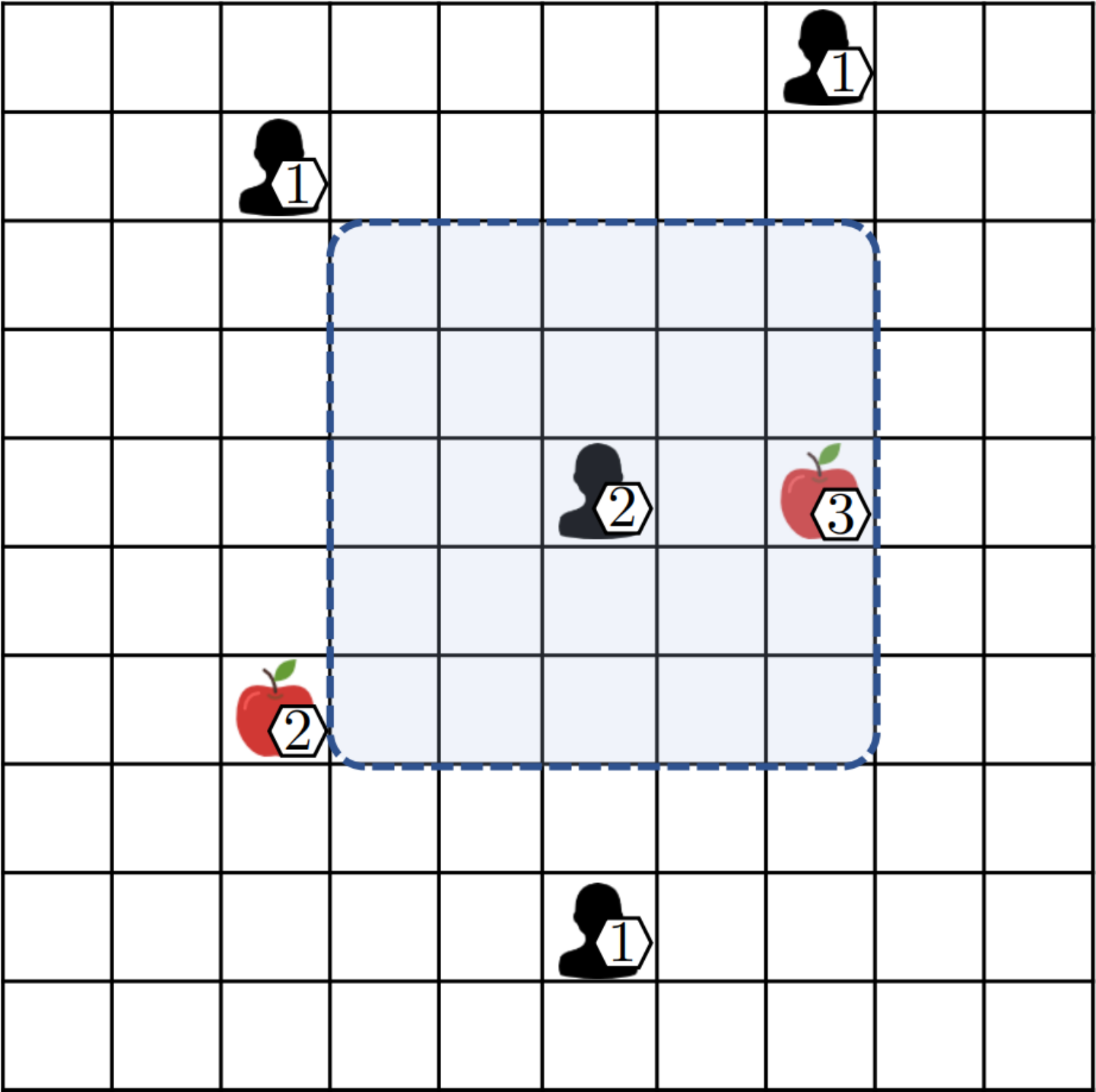}\label{fig:bench1}}
    	\subfigure[StarCraft Multi-Agent Challenge]{
		\includegraphics[width=0.33\linewidth]{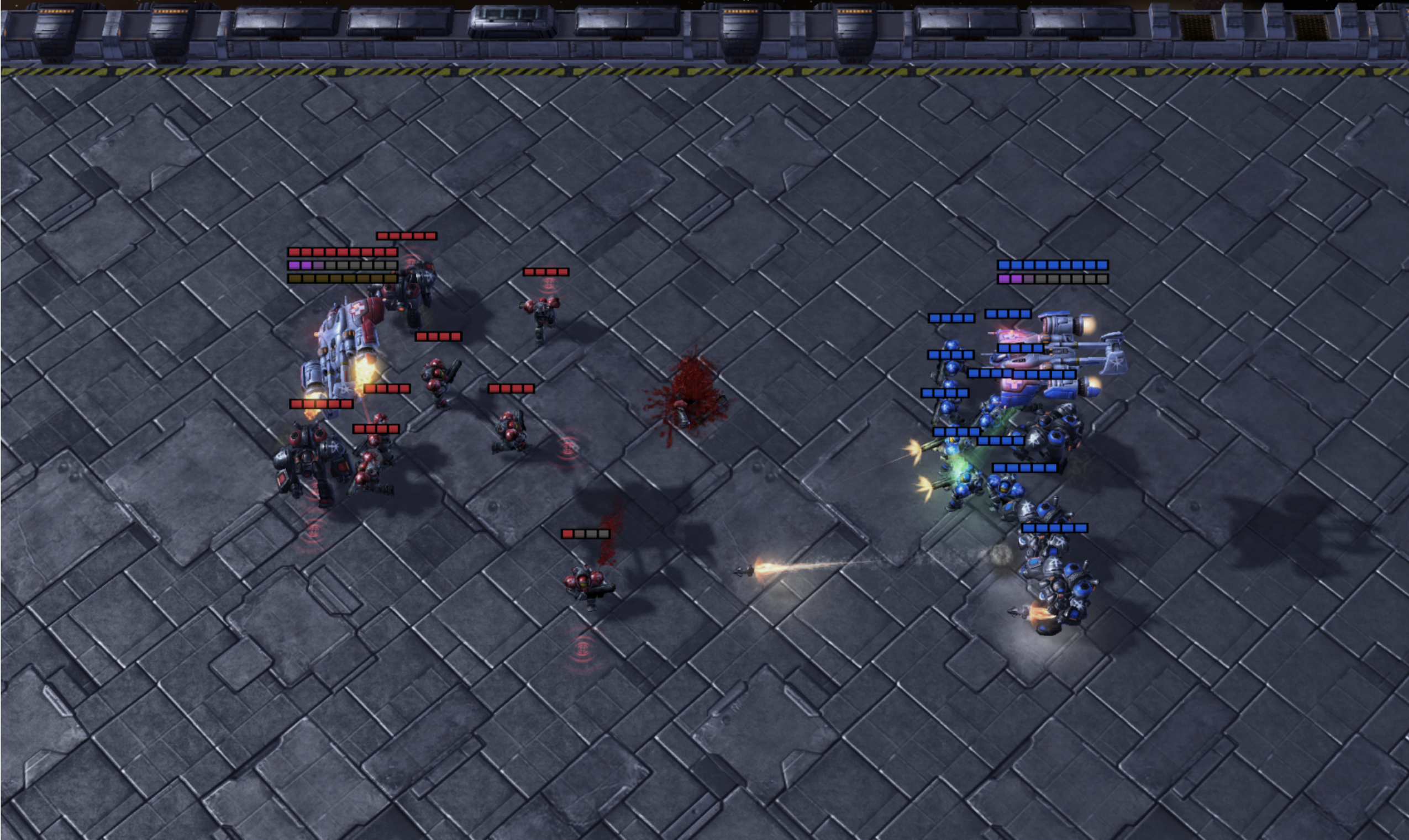}\label{fig:bench2}}
        \subfigure[StarCraft Multi-Agent Challenge v2]{
		\includegraphics[width=0.35\linewidth]{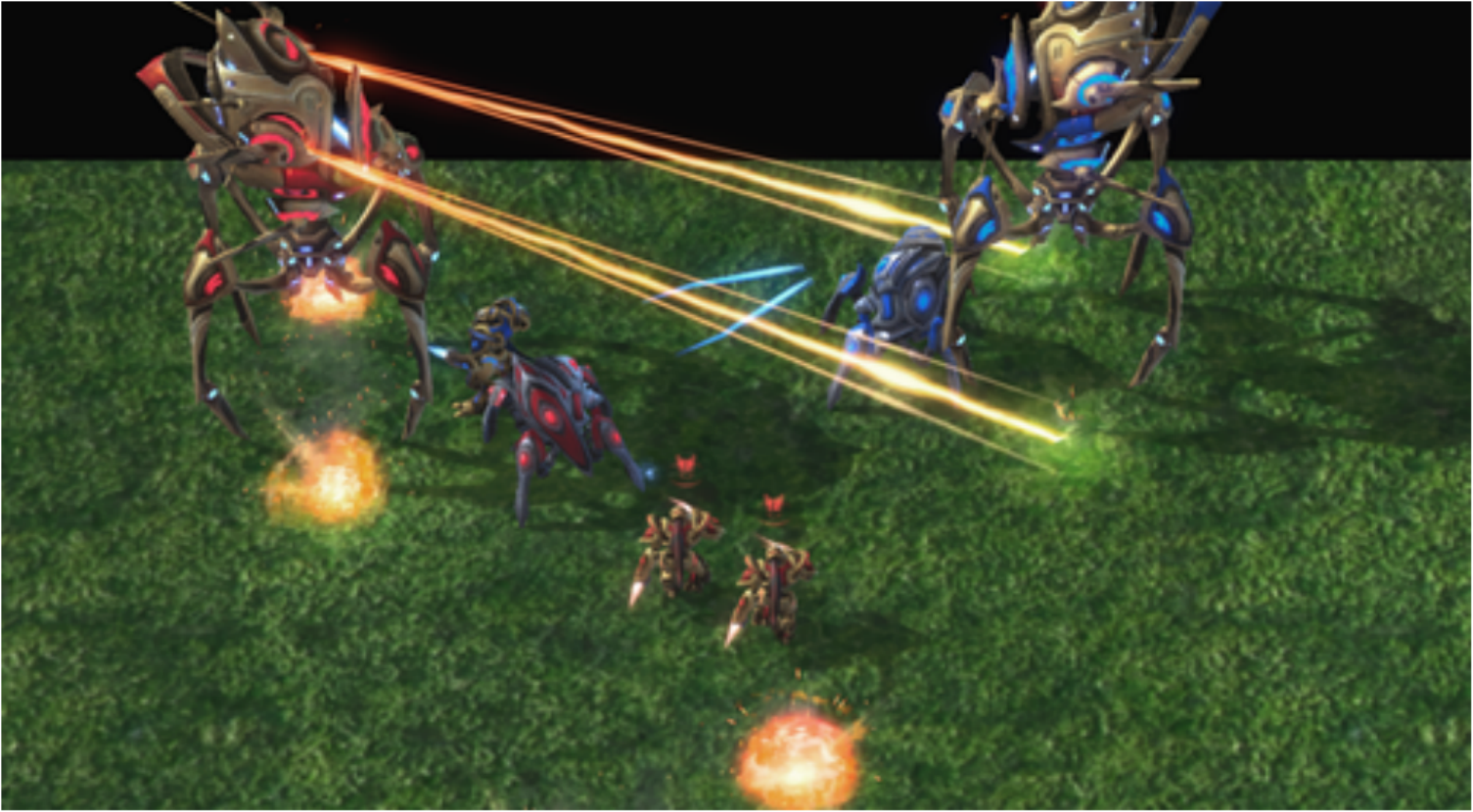}\label{fig:bench3}}
    \caption{Three benchmarks used in our experiments.}
    \label{fig:benchmark}
\end{figure*}

In our paper, we introduce three types of testing benchmarks as shown in Fig.~\ref{fig:benchmark}, including Level Based Foraging (LBF), StarCraft Multi-Agent Challenge (SMAC), and SMACv2. In this section, we describe the details and settings of these benchmarks.

\textbf{StarCraft Multi-Agent Challenge.} The StarCraft Multi-Agent Challenge (SMAC)~\cite{samvelyan2019starcraft} is a seminal benchmark for assessing the efficacy of multi-agent reinforcement learning (MARL) algorithms in challenging cooperative and competitive environments. SMAC focuses on micromanagement scenarios in StarCraft II (SC 2.4.10 version), where allied units are exclusively controlled by reinforcement learning agents, and enemy units are governed by a built-in AI with a fixed difficulty level of difficulty=7. Agents must collaboratively devise sophisticated strategies to outmaneuver and defeat the adversaries within a predefined exploration horizon. In this study, we evaluate algorithmic performance across 8 highly challenging combat scenarios, ensuring consistency and reproducibility through standardized training steps. Table~\ref{tab:smac} presents a brief introduction of these scenarios and the maximum training step, and all experimental configurations strictly adhere to the original setup, as detailed in \textit{Table~\ref{tab:sc2_settings}}.

\textbf{StarCraft Multi-Agent Challenge v2.} The StarCraft Multi-Agent Challenge v2 (SMACv2)~\cite{ellis2023smacv2} represents a significant evolution of the SMAC benchmark, designed to push the boundaries of MARL research by introducing heightened complexity and uncertainty. SMACv2 incorporates randomized initial configurations of enemy units, including their positions, attributes, and quantities, thus necessitating robust generalization capabilities. Furthermore, dynamic environmental features, such as map alterations and fluctuating objectives, create an additional layer of unpredictability. To ensure rigorous evaluation, we adopt uniform training step limits and leverage the SC2.4.10 environment version. In this work, we benchmark algorithms on 3 highly complex scenarios, with specific configurations outlined in \textit{Table~\ref{tab:smacv2}}, to comprehensively evaluate their adaptability and scalability.

\textbf{Level Based Foraging.} Level-Based Foraging (LBF)~\cite{NEURIPS2020_7967cc8e} is a well-established benchmark that encapsulates the interplay of cooperative and competitive dynamics in multi-agent systems. Agents are deployed on a $10 \times 10$ grid, each endowed with a specific level, and their observations are confined to a local $5 \times 5$ region surrounding them. The primary objective is to collaboratively consume food items scattered across the grid, contingent upon the collective level of participating agents meeting or exceeding the food item's level. Rewards are normalized based on the consumed food level, while a movement penalty of $-0.002$ incentivizes efficient exploration. This work considers two distinct task configurations: one comprising 4 agents and 2 food items, and another with 3 agents and 3 food items. Comprehensive experimental details are enumerated in \textit{Table~\ref{tab:lbf}}.

\begin{table}[htbp]\small
\centering
\caption{Introduction of scenarios in SMAC benchmark.}
\label{tab:smac}
\vspace{0.2cm}
\setlength{\tabcolsep}{2.5pt}
\begin{tabular}{lll@{}cc}
\toprule
\textsc{Map Name} & \textsc{Ally Units} & \textsc{Enemy Units} & \textsc{Total Timesteps} & \textsc{Scenario Type} \\
\midrule
\texttt{8m}       & \textsc{8 Marines}            & \textsc{8 Marines}            & \textsc{1M}                       & \textsc{Easy}                   \\
\texttt{2s3z}     & \textsc{2 Stalkers, 3 Zealots} & \textsc{2 Stalkers, 3 Zealots} & \textsc{1.5M}                      & \textsc{Easy}                   \\
\midrule
\texttt{3s5z}     & \textsc{3 Stalkers, 5 Zealots} & \textsc{3 Stalkers, 5 Zealots} & \textsc{1.5M}                      & \textsc{Hard}                   \\
\texttt{2c\_vs\_64zg} & \textsc{2 Colossi}          & \textsc{64 Zerglings}          & \textsc{1.5M}                      & \textsc{Hard}                   \\
\texttt{5m\_vs\_6m} & \textsc{5 Marines}           & \textsc{6 Marines}             & \textsc{1.5M}                      & \textsc{Hard}                   \\
\midrule
\texttt{3s5z\_vs\_3s6z} & \textsc{3 Stalkers, 5 Zealots} & \textsc{3 Stalkers, 6 Zealots} & \textsc{4M}                   & \textsc{Super Hard}             \\
\texttt{corridor} & \textsc{6 Zealots}             & \textsc{24 Zerglings}          & \textsc{4M}                      & \textsc{Super Hard}             \\
\texttt{6h\_vs\_8z} & \textsc{6 Hydralisks}         & \textsc{8 Zealots}             & \textsc{4M}                      & \textsc{Super Hard}             \\
\bottomrule
\end{tabular}
\end{table}

\begin{table}[htbp]\small
\centering
\caption{Introduction of scenarios in SMACv2 benchmark.}
\label{tab:smacv2}
\vspace{0.2cm}
\setlength{\tabcolsep}{2.5pt}
\begin{tabular}{l@{}ccl}
\toprule
\textsc{Map Name} & \textsc{Number of Allies} & \textsc{Number of Enemies} & \textsc{Unit Composition} \\
\midrule
\texttt{Terran\_10\_vs\_10}  & 10 & 10 & 
\textsc{Marines, Marauders, Medivacs} \\ 
\texttt{Zerg\_10\_vs\_10}    & 10 & 10 & 
\textsc{Zerglings, Hydralisks, Banelings} \\ 
\texttt{Protoss\_10\_vs\_10} & 10 & 10 &  
\textsc{Stalkers, Zealots, Colossi} \\ 
\bottomrule
\end{tabular}
\end{table}

\begin{table}[htbp]\small
\centering
\caption{Experimental Settings of Level-Based Foraging.}
\label{tab:lbf}
\vspace{0.2cm}
\begin{tabular}{lll}
\toprule
\textsc{Hyperparameter}        & \textsc{Value}       & \textsc{Description}                           \\ 
\midrule
\textsc{Max Player Level}      & 3                   & \textsc{Maximum Agent Level Attribute}        \\
\textsc{Max Episode Length}    & 50                  & \textsc{Maximum Timesteps Per Episode}        \\
\textsc{Batch Size}            & 32                  & \textsc{Number of Episodes Per Update}        \\
\textsc{Test Interval}         & 10,000              & \textsc{Frequency of Evaluating Performance}  \\
\textsc{Test Episodes}         & 32                  & \textsc{Number of Episodes to Test}           \\
\textsc{Replay Batch Size}     & 5000                & \textsc{Maximum Number of Episodes Stored in Memory} \\
\textsc{Discount Factor $\gamma$} & 0.99            & \textsc{Degree of Impact of Future Rewards}   \\
\textsc{Total Timesteps}       & 1,050,000           & \textsc{Number of Training Steps}             \\
\textsc{Start $\epsilon$}      & 1.0                 & \textsc{The Start $\epsilon$ Value to Explore} \\
\textsc{Finish $\epsilon$}     & 0.05                & \textsc{The Finish $\epsilon$ Value to Explore} \\
\textsc{Anneal Steps for $\epsilon$} & 50,000        & \textsc{Number of Steps of Linear Annealing}  \\
\textsc{Target Update Interval} & 200                & \textsc{The Target Network Update Cycle}      \\
\bottomrule
\end{tabular}
\end{table}

\begin{table}[htbp]\small
\centering
\caption{Experimental settings of SMAC and SMACv2.}
\label{tab:sc2_settings}
\vspace{0.2cm}
\begin{tabular}{lcl}
\toprule
\textsc{Hyperparameter}               & \textsc{Value}   & \textsc{Description}                                 \\
\midrule
\textsc{Difficulty}                   & \textsc{7}       & \textsc{Enemy units with built-in AI difficulty}     \\
\textsc{Batch Size (SMAC)}                   & \textsc{32}      & \textsc{Number of episodes per update}              \\
\textsc{Batch Size (SMACv2)}                   & \textsc{128}      & \textsc{Number of episodes per update}              \\
\textsc{Test Interval}                & \textsc{10,000}  & \textsc{Frequency of evaluating performance}         \\
\textsc{Test Episodes}                & \textsc{32}      & \textsc{Number of episodes to test}                 \\
\textsc{Replay Batch Size}            & \textsc{5000}    & \textsc{Maximum number of episodes in memory} \\
\textsc{Discount Factor} $\gamma$     & \textsc{0.99}    & \textsc{Degree of impact of future rewards}          \\
\textsc{Start} $\varepsilon$          & \textsc{1.0}     & \textsc{The start $\varepsilon$ value to explore}    \\
\textsc{Finish} $\varepsilon$         & \textsc{0.05}    & \textsc{The finish $\varepsilon$ value to explore}   \\
\textsc{Anneal Steps for Easy \& Hard}& \textsc{50,000}  & \textsc{Number of steps of linear annealing $\varepsilon$} \\
\textsc{Anneal Steps for Super Hard}  & \textsc{100,000} & \textsc{Number of steps of linear annealing $\varepsilon$} \\
\textsc{Target Update Interval}       & \textsc{200}     & \textsc{The target network update cycle}             \\
\bottomrule
\end{tabular}
\end{table}

\subsection{Hyperparameters of Baselines}\label{appendix2}
Our experiments are conducted using the PyMARL framework~\cite{samvelyan2019starcraft}, which we adopt to implement a range of baseline methods. These include five representative value-decomposition algorithms—VDN~\cite{vdn}, QMIX~\cite{qmix}, QTRAN~\cite{qtran}, QPLEX~\cite{qplex}, and CDS~\cite{li2021celebrating}—as well as four subtask-oriented approaches: RODE~\cite{wang2021rode}, GoMARL~\cite{zang2023automatic}, ACORM~\cite{hu2024attention}, and DT2GS~\cite{tian2023decompose}.
For fair comparison, all baselines are configured with consistent hyperparameters, following the default settings provided in the official PyMARL codebases and respective implementations. 
Our method is trained with evaluations every $20,000$ environment steps, using $20$ episodes and decentralized greedy action selection to estimate the test win rate.
To ensure statistical robustness, we run all experiments with five random seeds in environments characterized by clear success/failure dynamics.
The win rate is computed as the proportion of episodes in which the allied agents successfully eliminate all enemy units within the time limit.
Optimization is performed using the RMSprop optimizer with a learning rate of $0.0005$, and target network updates are conducted every $200$ training episodes.

\subsection{Hyperparameters of C$\text{D}^\text{3}$T}

Our implementation of C$\text{D}^\text{3}$T is built upon the PyMARL framework. Agent observations are manually encoded and projected into a 32-dimensional hidden space.
Each agent's policy network comprises an MLP layer followed by a GRU layer, with the final embedding dimension set to 64.
All hyperparameters used in C$\text{D}^\text{3}$T are summarized in \textit{Table~\ref{tab:cd3t_hyperparams}}. 
Training is performed on a single machine equipped with an NVIDIA RTX 4090 GPU and an Intel Core i9-13900K CPU. 
The total training time varies depending on the complexity of the scenario.

\begin{table}[htbp]
\centering
\caption{The special hyper-parameters of C$\text{D}^\text{3}$T architecture.}
\vspace{0.2cm}
\begin{tabular}{lccp{8cm}}
\toprule
\textsc{Component} & \textsc{Hyper-parameters} & \textsc{Value} \\ 
\midrule
\multirow{4}{*}{\textsc{Diffusion Model}} 
& \textsc{Start} $\beta$ & 0.0001 \\ 
& \textsc{End} $\beta$ & 0.02 \\ 
& \textsc{Scaling Factor} $\lambda_{dr}$ & 10  \\ 
& \textsc{Scaling Factor} $\eta_e$ & 0.1  \\ 
\midrule
\multirow{5}{*}{\textsc{Subtask Selector}} 
& \textsc{Number of Subtask Clusters} & 3 \\ 
& \textsc{Subtask Selection Interval} & 5  \\ 
& \textsc{State Latent Dimension} & 32  \\ 
& \textsc{Action Latent Dimension} & 20 \\ 
& \textsc{Subtask Action Spaces Update Start} & 50,000  \\ 
\midrule
\multirow{7}{*}{\textsc{Mixing Network}} 
& \textsc{State Embedding Dimension} & 32 \\ 
& \textsc{Number of Attention Heads} & 8 \\ 
& \textsc{First Layer Query Dimension} & 64 \\ 
& \textsc{Second Layer Query Dimension} & 32 \\ 
& \textsc{Layer Key Dimension} & 32 \\ 
& \textsc{First Layer Weight Dimension} & 64 \\ 
& \textsc{Second Layer Weight Dimension} & 8 \\ 
\bottomrule
\label{tab:cd3t_hyperparams}
\end{tabular}
\end{table}

\appsection{Performance on SMACv2}\label{exp_smacv2}

\begin{figure*}[h]
\begin{center}
    \includegraphics[width=\textwidth]{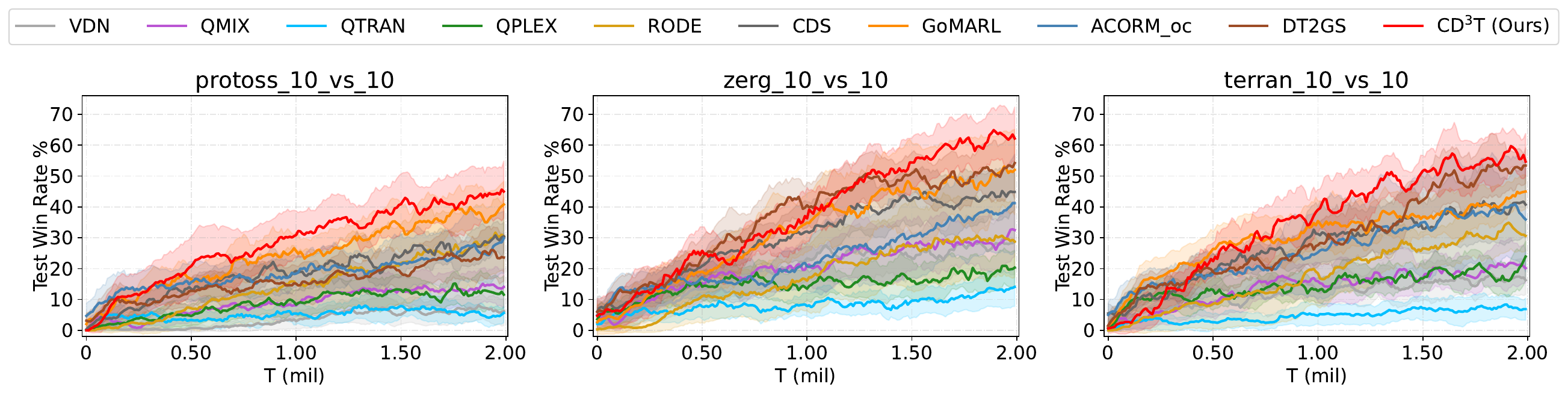}
\end{center}
\vspace{-0.3cm}
\caption{Performance comparison with baselines on SMACv2 scenarios.}
\label{fig:smacv2}
\end{figure*}

The introduction of SMACv2 was motivated by the observation that many original SMAC maps lack sufficient diversity and stochasticity to effectively challenge state-of-the-art MARL algorithms.
To evaluate the generalization and robustness of our proposed method under more complex and less predictable conditions, we benchmark C$\text{D}^3$T against several baselines on three redesigned scenarios in SMACv2: \texttt{protoss\_10\_vs\_10}, \texttt{zerg\_10\_vs\_10}, and \texttt{terran\_10\_vs\_10}.

Figure~\ref{fig:smacv2} presents the comparative performance across these tasks.
C$\text{D}^3$T achieves better results, suggesting enhanced adaptability in handling diverse and intricate coordination challenges. 
In contrast, traditional value factorization methods such as VDN, QMIX, QTRAN, and QPLEX exhibit notably weaker performance.
This may reflect their limitations in modeling complex interactions, particularly due to rigid decomposition structures that often conflate individual utility with joint return signals.
RODE performs less effectively, possibly due to its simplified semantic extraction mechanism and limited propagation of semantic context through the learning pipeline.
The lack of fine-grained semantic representations may hinder accurate credit assignment under decentralized settings.
CDS achieves reasonable performance, likely benefiting from its emphasis on promoting diversity among agents, which can be advantageous in dynamic environments with shifting coordination demands.
GoMARL outperforms traditional baselines by leveraging group-level semantics to enrich the value decomposition process. 
ACORM also demonstrates relatively competitive results, indicating that its structure remains effective within the 2M timestep horizon, even without explicit clustering mechanisms.
DT2GS shows performance close to that of C$\text{D}^3$T on \texttt{zerg\_10\_vs\_10} and \texttt{terran\_10\_vs\_10}, but underperforms on the more demanding \texttt{protoss\_10\_vs\_10} task.
This gap suggests that while DT2GS captures some useful subtask structure, its strong reliance on subtask transferability may limit its ability to adapt to task-specific complexities.
Overall, these empirical results indicate that C$\text{D}^3$T is a promising method for dynamically decomposing and coordinating multi-agent tasks, which should be attributed to its use of conditional diffusion modeling and latent representation guidance in value factorization.

\begin{table}[htbp]
\centering
\caption{Scalability comparison across QMIX, RODE, CD$^3$T and CD$^3$T (After $50K$ timesteps).}
\vspace{0.2cm}
\setlength{\tabcolsep}{6pt}
\begin{tabular}{lcccc}
\toprule
\textsc{Component} & \textsc{QMIX} & \textsc{RODE} & \textsc{CD$^3$T} & \textsc{CD$^3$T (After $50K$ timesteps)} \\
\midrule
\textsc{Agent Network} & 27.5K & 27.1K & 27.1K & 27.1K \\
\textsc{Mixing Network} & 18.8K & 18.8K & 23.2K & 23.2K \\
\textsc{Diffusion Encoder} & -- & -- & 55.0K & -- \\
\textsc{Other Modules} & -- & 25.4K & 22.9K & 22.9K \\
\midrule
\textsc{Total Parameters} & \textbf{46.3K} & \textbf{71.3K} & \textbf{128.2K} & \textbf{73.2K} \\
\bottomrule
\label{tab:compare_qmix_rode_cd3t_clean}
\end{tabular}
\vspace{-0.2cm}
\end{table}

\subsection{Model Scalability and Training Efficiency Analysis}

To reduce memory overhead and improve training efficiency, we adopt a lightweight diffusion model, with detailed parameter statistics listed in Table~\ref{tab:compare_qmix_rode_cd3t_clean}. 
After the initial $50\text{K}$ timesteps, subtask assignments and their corresponding latent representations are fixed, eliminating the need to invoke the diffusion model for subsequent action representation generation. 
At this point, the overall parameter count of C$\text{D}^3$T becomes comparable to that of RODE.
Since the first $50\text{K}$ timesteps constitute only a small portion of the total training horizon, the end-to-end training time of C$\text{D}^3$T remains similar to RODE—typically ranging from 1 to 4 million timesteps and requiring approximately 6 to 25 hours of wall-clock time, depending on scenario complexity.
Despite operating under a comparable computational budget, C$\text{D}^3$T consistently outperforms all baselines across nearly all evaluated scenarios, highlighting its empirical effectiveness and scalability in diverse multi-agent tasks.

\appsection{Related Work}\label{appendix}
\textbf{Value-based MARL.} Value-based MARL (Multi-Agent Reinforcement Learning) has witnessed significant advancements in recent years. 
A basic approach is Independent Q-Learning (IQL), which treats other agents as part of a non-stationary environment and trains separate Q-functions for each agent.
Despite its simplicity, IQL often suffers from instability due to policy shifts among agents, leading to poor convergence.
At the other extreme, centralized methods learn a joint Q-function over the full action-observation space, effectively capturing inter-agent dependencies.
However, this comes at the cost of scalability, as the joint space grows exponentially with the number of agents.

To balance scalability and coordination, many recent methods adopt centralized training with decentralized execution (CTDE) via value function factorization. 
VDN~\cite{vdn} introduces a linear additive decomposition of the global Q-function. 
QMIX~\cite{qmix} extends this with a monotonic mixing network, ensuring consistency between local and global values. 
QTRAN~\cite{qtran} offers a more flexible formulation with theoretical guarantees, while QPLEX~\cite{qplex} incorporates the Individual-Global-Max (IGM) principle into a duplex dueling architecture, enhancing expressiveness and learning efficiency.
Improving credit assignment remains another key focus. 
CDS~\cite{li2021celebrating} employs counterfactual baselines and agent-specific utilities to better align local contributions with global objectives. 
GoMARL~\cite{zang2023automatic} introduces group-wise factorization, partitioning agents into subsets to reduce the joint action space and improve scalability in large-scale scenarios.
Beyond factorization, complementary advances span differentiable communication~\cite{wang2020learning}, exploration strategies~\cite{gupta2021uneven}, and robustness under non-stationarity~\cite{zhang2020robust}.

Building on these advancements, our method C$\text{D}^3$T introduces a hierarchical framework that leverages diffusion-based latent representations for subtask discovery. 
These representations are fed into an attention-based mixing network to enable semantically guided and interpretable credit assignment, offering improved scalability and adaptability in complex MARL settings.

\noindent\textbf{Hierarchical MARL.} Hierarchical reinforcement learning (HRL) has been extensively studied in single-agent settings to address sparse rewards and facilitate transfer learning~\cite{al2015hierarchical,sutton1999between}. 
Existing works in HRL typically focus on decomposing tasks through either subgoal learning\cite{nair2020hierarchical,dilokthanakul2019feature} or skill discovery~\cite{daniel2016hierarchical,sharma2020dynamics}.

Extending HRL to multi-agent scenarios introduces additional challenges, such as scalable communication and effective role differentiation~\cite{wang2020roma}, which call for hierarchical coordination among agents~\cite{zhang2010self}.
Early work, such as feudal RL~\cite{dayan1992feudal}, adopts a manager-worker paradigm, where high-level managers assign goals and lower-level workers act to fulfill them.
Bi-level hierarchies have also been successfully applied to discover agent-level skills~\cite{lee2020learning,yang2020hierarchical}.
Recent efforts have extended hierarchical MARL to Markov games, where high-level policies govern strategic choices across agents~\cite{wang2021rode}.
In particular, HAVEN~\cite{xu2023haven} introduces a hierarchical value decomposition framework with dual coordination mechanisms—across levels and across agents—enabling flexible cooperation without the need for domain-specific priors or pretraining.
By combining structured credit assignment with hierarchical reasoning, HAVEN improves coordination in complex tasks.

Here, C$\text{D}^\text{3}$T introduces a novel two-level hierarchical architecture that automatically identifies subtask structures and optimizes policies within each subspace.
This design improves credit assignment, enhances sample efficiency, and enables robust performance in dynamic, high-dimensional multi-agent scenarios.

\noindent\textbf{Diffusion Models for Reinforcement Learning.} Diffusion models have recently emerged as a powerful class of generative frameworks, leveraging denoising processes to reverse multi-step noise corruptions and generate high-fidelity samples~\cite{song2021score}.
Their strong generative capacity has sparked growing interest in reinforcement learning (RL), leading to a range of innovative applications.
A representative example is Diffuser~\cite{janner2022planning}, which learns a diffusion-based model from offline trajectories to perform goal-directed planning via guided sampling.
Beyond trajectory generation, diffusion models have shown promise across multiple RL components, including serving as expressive policy parameterizations~\cite{wang2023diffusion}, enabling structured exploration through latent skill modeling~\cite{venkatraman2024reasoning}, and augmenting experience data to improve learning stability~\cite{lu2023synthetic}.
They have also been increasingly applied in high-level decision-making, spanning tasks such as multi-task RL~\cite{he2023diffusion}, imitation learning~\cite{hegde2023generating}, and motion trajectory synthesis~\cite{zhang2024motiondiffuse}.
C$\text{D}^3$T builds on these advances by leveraging diffusion models to learn expressive latent representations that capture complex distributional properties, such as skewness and multi-modality—essential for reliable subtask discovery. 
This enables a more principled and flexible approach to hierarchical decision-making in cooperative MARL.

\begin{figure}[htbp]
  \centering

  \begin{minipage}{\textwidth}
    \centering
    \includegraphics[width=0.24\textwidth]{vis1}
    \includegraphics[width=0.24\textwidth]{vis2}
    \includegraphics[width=0.24\textwidth]{vis3}
    \includegraphics[width=0.24\textwidth]{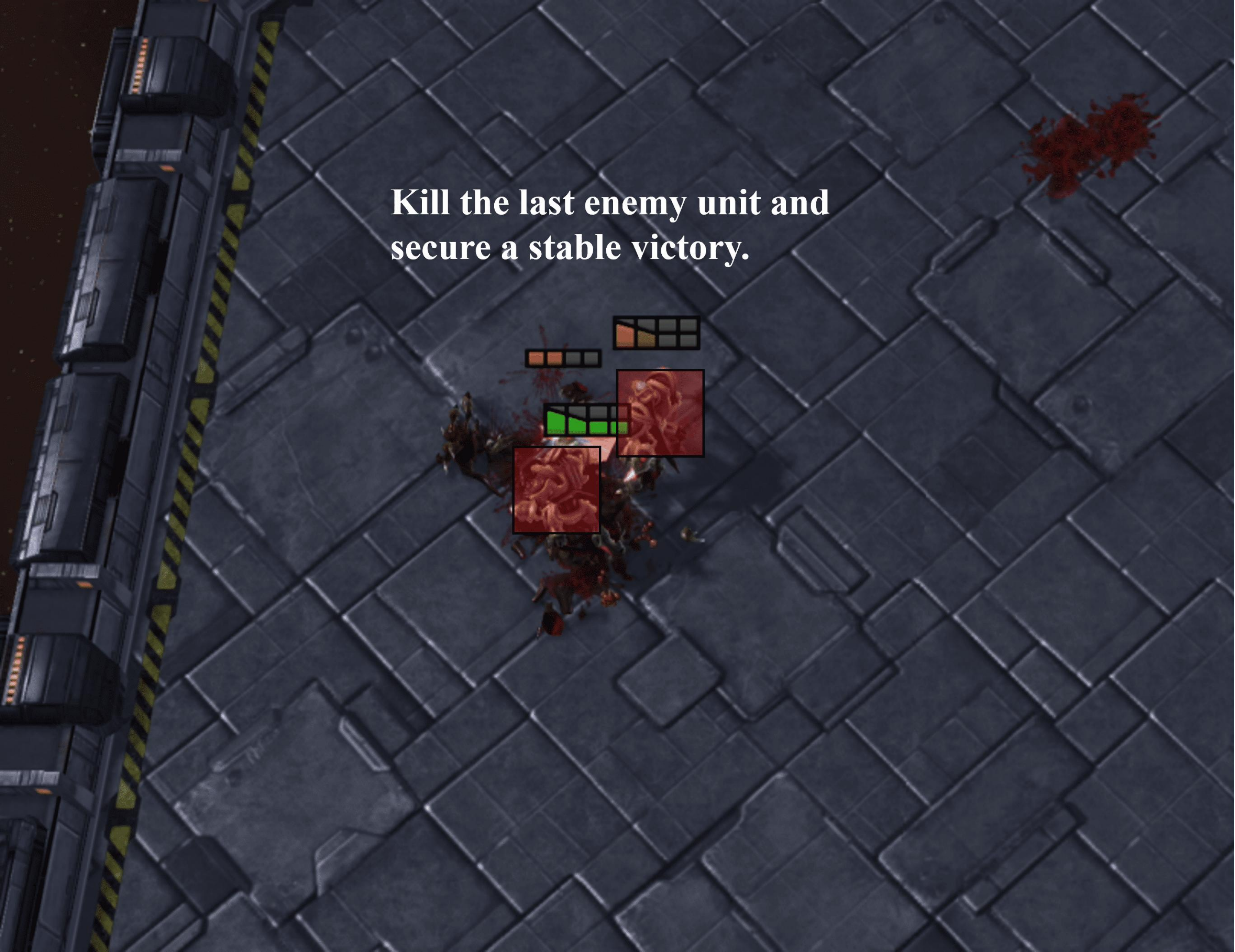}
    \\ \vspace{1mm}
    (a) Visualizations of CD$^3$T in one episode on \texttt{corridor}.
  \end{minipage}

  \vspace{3mm}

  \begin{minipage}{\textwidth}
    \centering
    \includegraphics[width=0.24\textwidth]{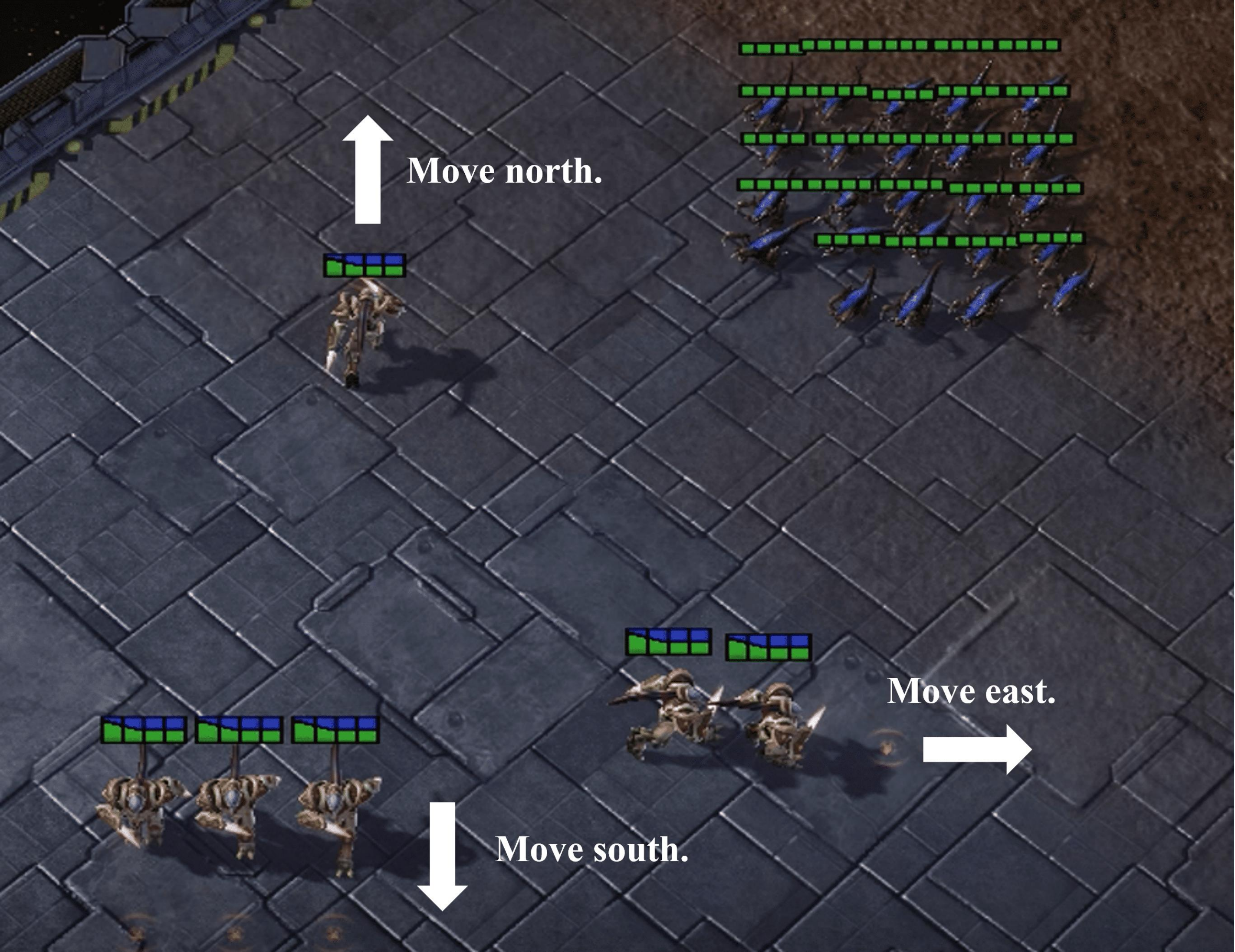}
    \includegraphics[width=0.24\textwidth]{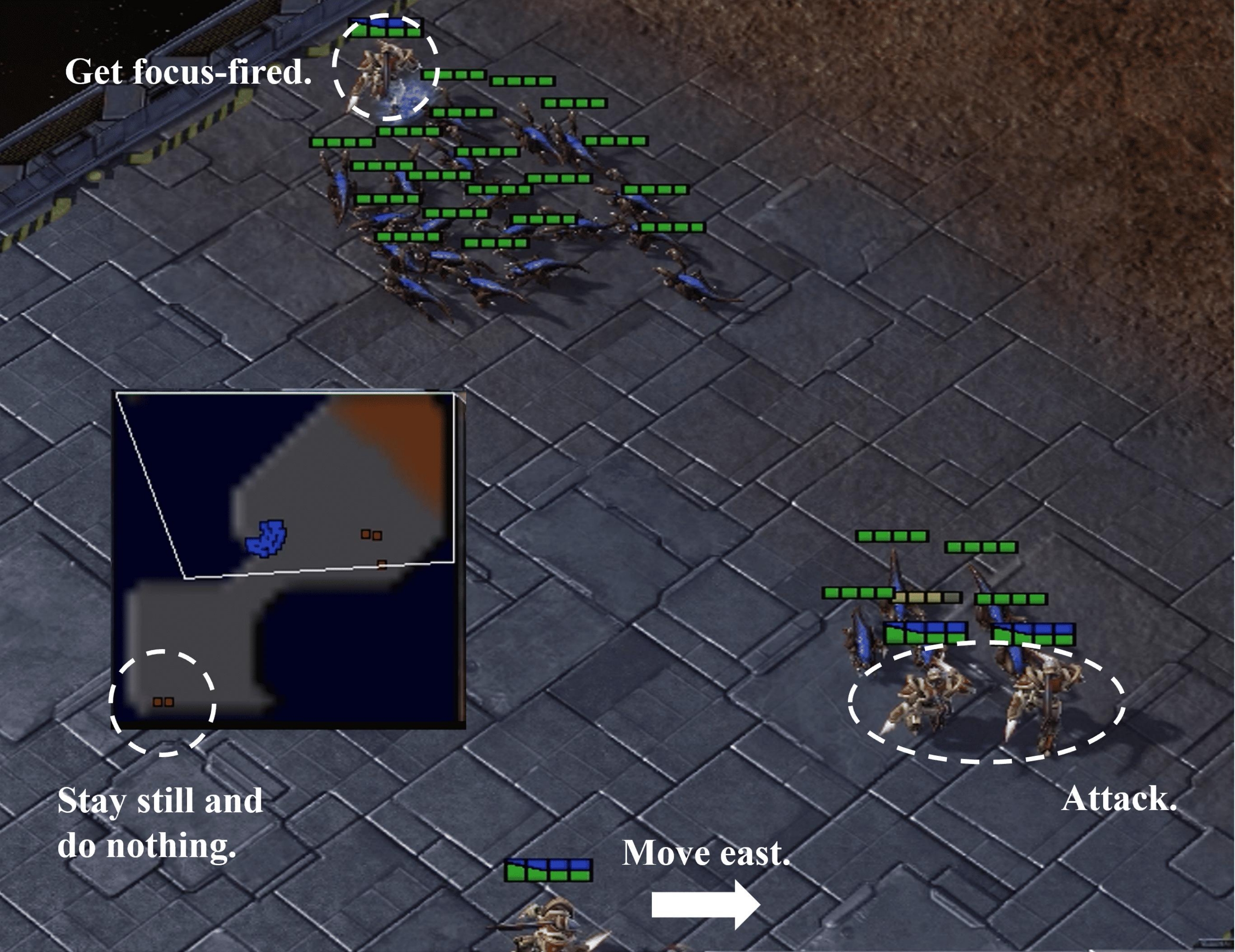}
    \includegraphics[width=0.24\textwidth]{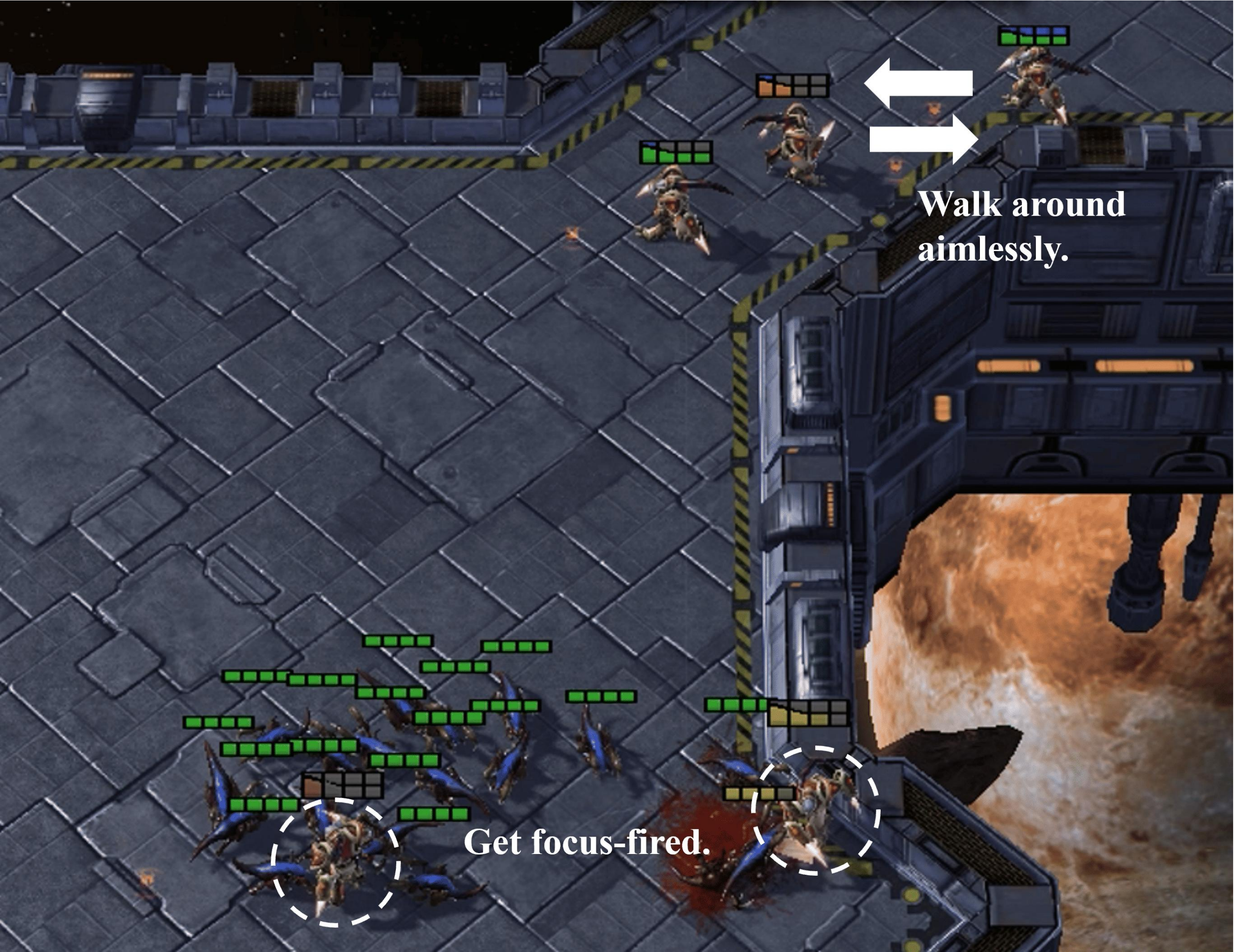}
    \includegraphics[width=0.24\textwidth]{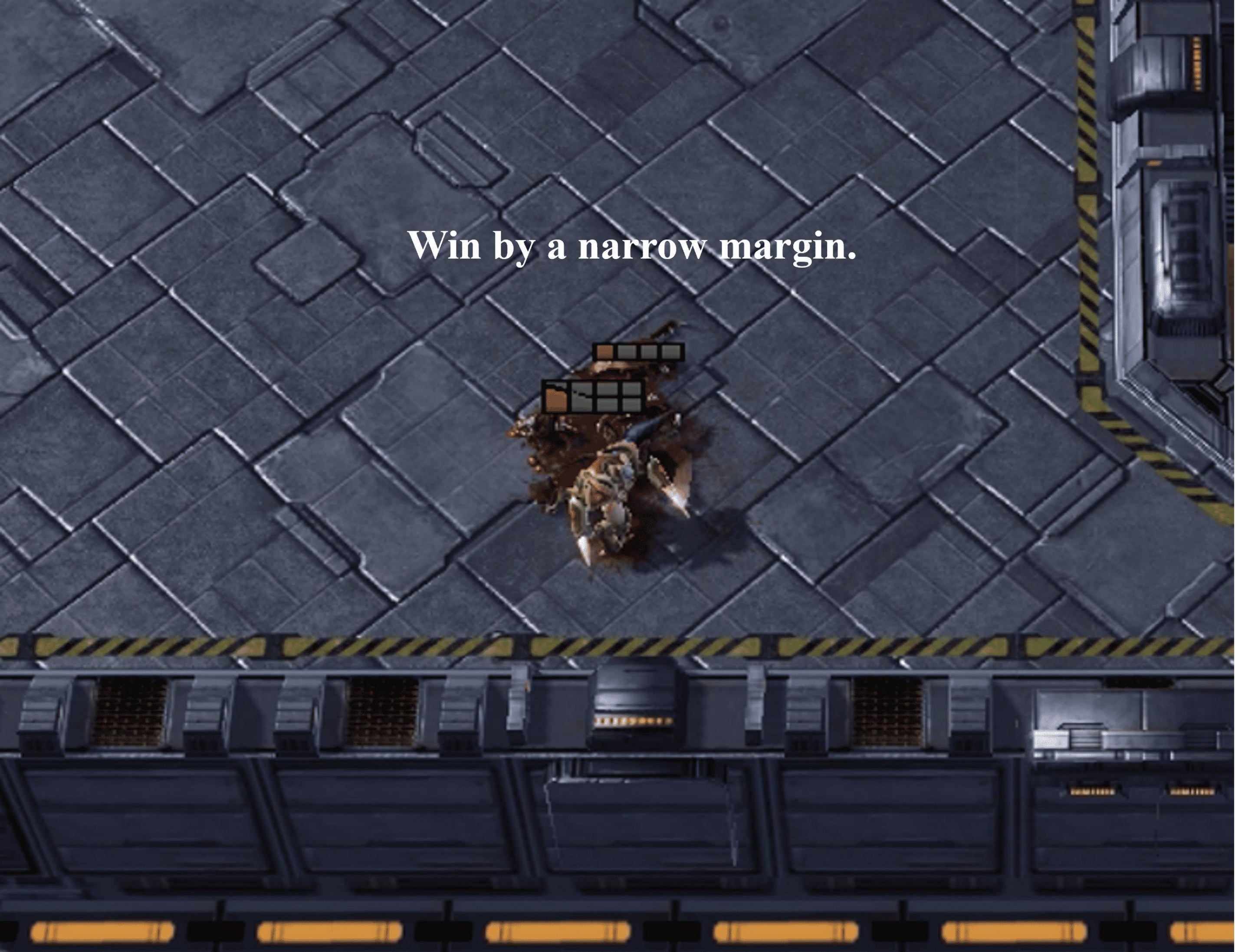}
    \\ \vspace{1mm}
    (b) Visualizations of RODE in one episode on \texttt{corridor}.
  \end{minipage}

  \vspace{3mm}

  \begin{minipage}{\textwidth}
    \centering
    \includegraphics[width=0.24\textwidth]{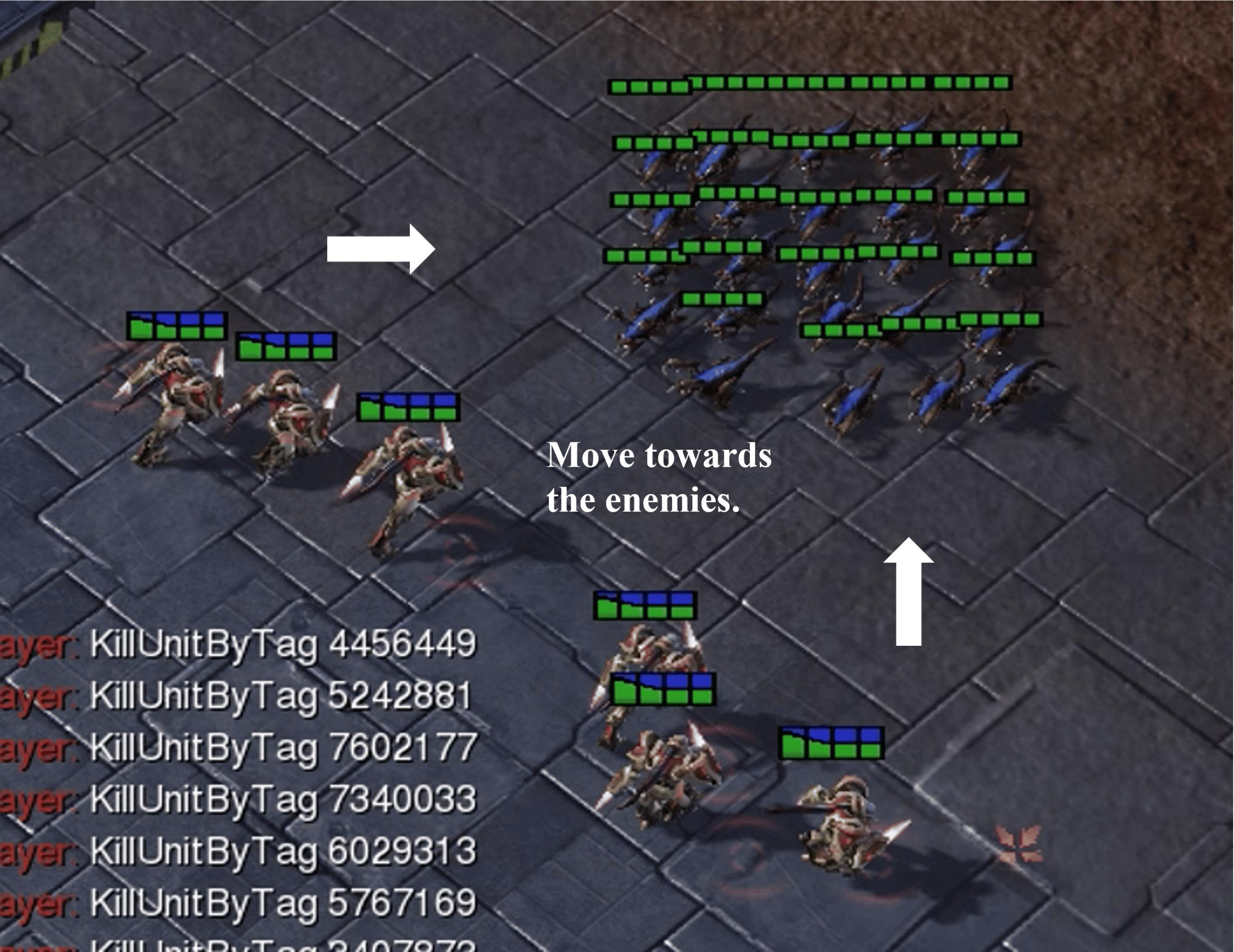}
    \includegraphics[width=0.24\textwidth]{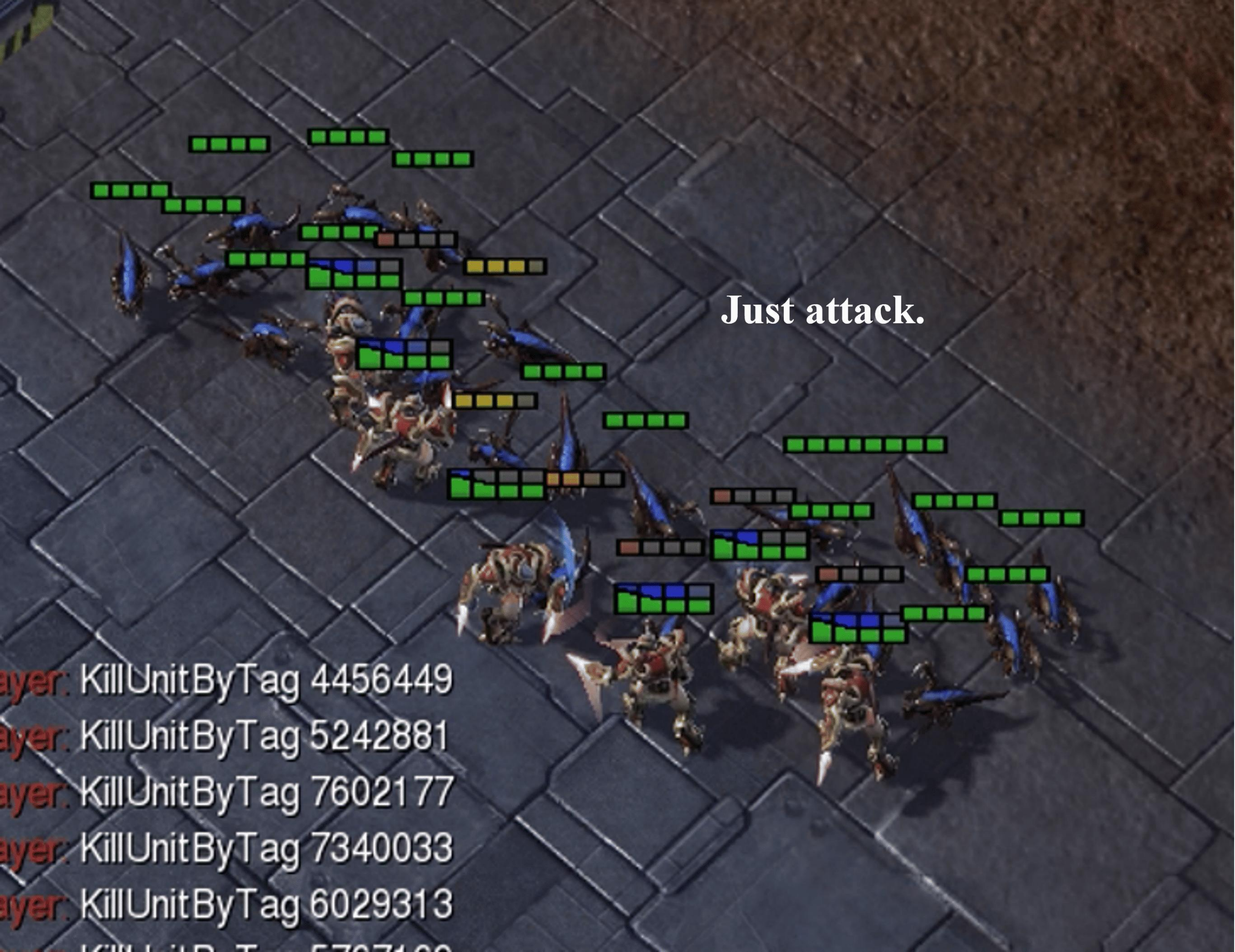}
    \includegraphics[width=0.24\textwidth]{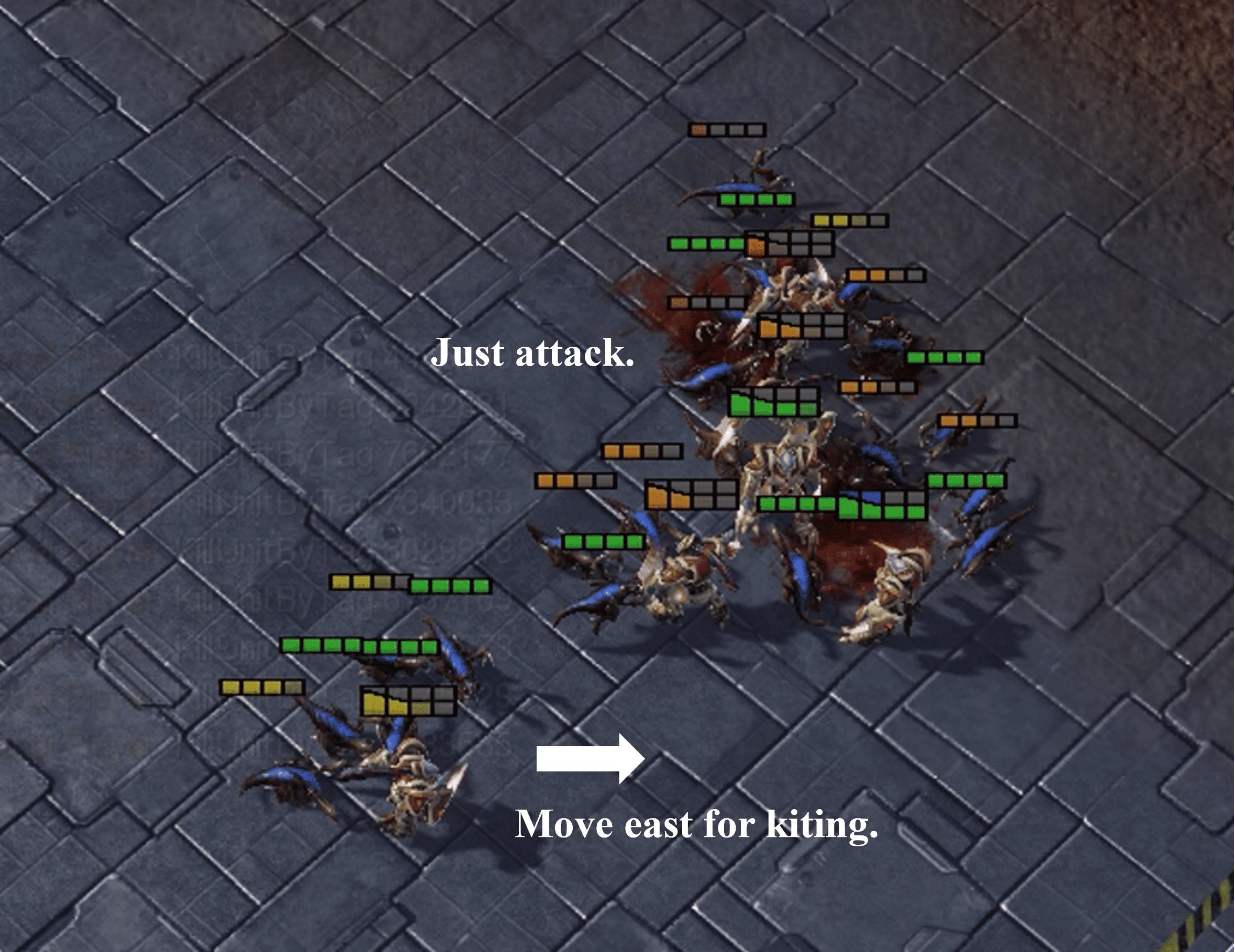}
    \includegraphics[width=0.24\textwidth]{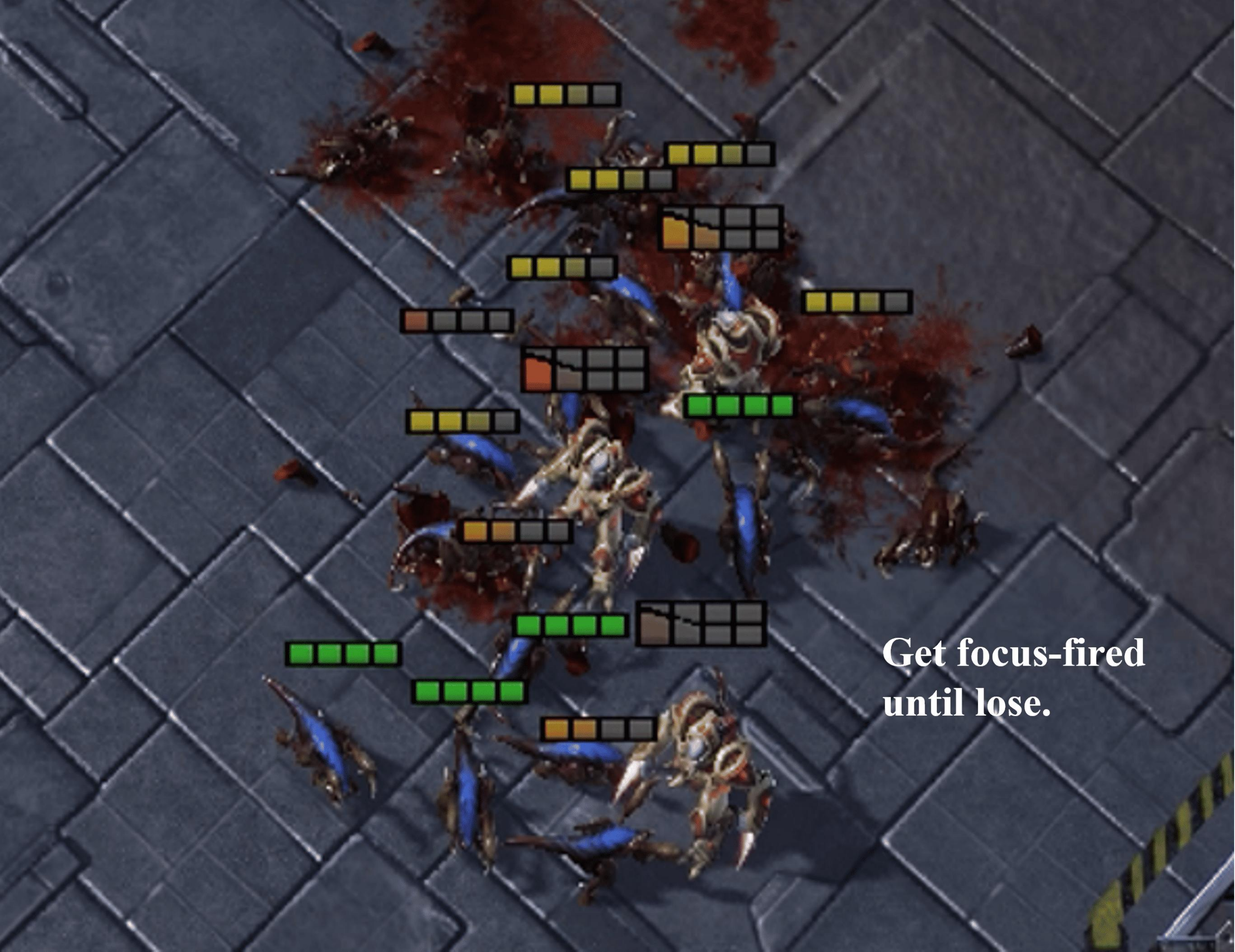}
    \\ \vspace{1mm}
    (c) Visualizations of QMIX in one episode on \texttt{corridor}.
  \end{minipage}

    \caption{Qualitative comparison of CD$^3$T, RODE, and QMIX on \texttt{corridor}. Each row depicts a sequence of visualizations from a single episode under the same scenario, illustrating the agent behaviors and coordination patterns exhibited by each method.}
    \label{fig:vis_comparison}
\end{figure}

\appsection{Visulizations}\label{appendix:vis}

\subsection{Visual Evidence of Effective Task Decomposition}
To further illustrate the reasonableness of CD$^3$T in task decomposition and subtask allocation, we visualize agent trajectories and subtask assignments on the \texttt{corridor} scenario across two representative methods. 
As shown in \textit{Fig.~\ref{fig:vis_comparison}(a)}, C$\text{D}^3$T maintains a consistently high win rate, underscoring the benefits of its structured decision-making (see Section~\ref{sec:vis}). 
Even in the late stage of combat, when only two of six allied agents remain, one agent continues to operate strategically with considerable remaining health.
This indicates that the subtask representations learned by C$\text{D}^3$T enable robust and context-aware behavior, even under significant pressure of being killed.
In contrast, \textit{Fig.~\ref{fig:vis_comparison}(b)} reveals clear deficiencies in RODE’s coordination capabilities. 
Three allied agents act ineffectively from the beginning, with two retreating to a corner and remaining idle for most of the episode. 
As the remaining agents come under concentrated enemy fire, they display hesitant and uncoordinated movement, failing to provide support or leverage their numerical advantage. 
As a result, RODE secures only a marginal win, despite an initially favorable position.
\textit{Fig.~\ref{fig:vis_comparison}(c)} further illustrates the limitations of QMIX, which lacks any discernible tactical structure.
 Agents engage in an uncoordinated frontal assault and are rapidly eliminated by focused enemy attacks.
 Although one agent sporadically exhibits basic kiting behavior, the absence of a collective strategy renders such efforts ineffective.

These visualizations highlight the advantage of C$\text{D}^3$T's hierarchical framework, which facilitates clear subtask delineation and consistent coordination. 
By leveraging diffusion-based representations, agents are able to adapt their roles dynamically and cooperate efficiently, resulting in resilient and strategically sound group behavior.

\newpage
\setlength{\leftmargini}{20pt}
\makeatletter\def\@listi{\leftmargin\leftmargini \topsep .5em \parsep .5em \itemsep .5em}
\def\@listii{\leftmargin\leftmarginii \labelwidth\leftmarginii \advance\labelwidth-\labelsep \topsep .4em \parsep .4em \itemsep .4em}
\def\@listiii{\leftmargin\leftmarginiii \labelwidth\leftmarginiii \advance\labelwidth-\labelsep \topsep .4em \parsep .4em \itemsep .4em}\makeatother

\setcounter{secnumdepth}{0}
\renewcommand\thesubsection{\arabic{subsection}}
\renewcommand\labelenumi{\thesubsection.\arabic{enumi}}

\newcounter{checksubsection}
\newcounter{checkitem}[checksubsection]

\newcommand{\checksubsection}[1]{%
  \refstepcounter{checksubsection}%
  \paragraph{\arabic{checksubsection}. #1}%
  \setcounter{checkitem}{0}%
}

\newcommand{\checkitem}{%
  \refstepcounter{checkitem}%
  \item[\arabic{checksubsection}.\arabic{checkitem}.]%
}
\newcommand{\question}[2]{\normalcolor\checkitem #1 #2 \color{blue}}
\newcommand{\ifyespoints}[1]{\makebox[0pt][l]{\hspace{-15pt}\normalcolor #1}}

\section*{Reproducibility Checklist}


\checksubsection{General Paper Structure}
\begin{itemize}

\question{Includes a conceptual outline and/or pseudocode description of AI methods introduced}{(yes/partial/no/NA)}
yes

\question{Clearly delineates statements that are opinions, hypothesis, and speculation from objective facts and results}{(yes/no)}
yes

\question{Provides well-marked pedagogical references for less-familiar readers to gain background necessary to replicate the paper}{(yes/no)}
yes

\end{itemize}
\checksubsection{Theoretical Contributions}
\begin{itemize}

\question{Does this paper make theoretical contributions?}{(yes/no)}
yes

	\ifyespoints{\vspace{1.2em}If yes, please address the following points:}
        \begin{itemize}
	
	\question{All assumptions and restrictions are stated clearly and formally}{(yes/partial/no)}
	yes

	\question{All novel claims are stated formally (e.g., in theorem statements)}{(yes/partial/no)}
	yes

	\question{Proofs of all novel claims are included}{(yes/partial/no)}
	yes

	\question{Proof sketches or intuitions are given for complex and/or novel results}{(yes/partial/no)}
	yes

	\question{Appropriate citations to theoretical tools used are given}{(yes/partial/no)}
	yes

	\question{All theoretical claims are demonstrated empirically to hold}{(yes/partial/no/NA)}
	yes

	\question{All experimental code used to eliminate or disprove claims is included}{(yes/no/NA)}
	NA
	
	\end{itemize}
\end{itemize}

\checksubsection{Dataset Usage}
\begin{itemize}

\question{Does this paper rely on one or more datasets?}{(yes/no)}
no

\ifyespoints{If yes, please address the following points:}
\begin{itemize}

	\question{A motivation is given for why the experiments are conducted on the selected datasets}{(yes/partial/no/NA)}
	NA

	\question{All novel datasets introduced in this paper are included in a data appendix}{(yes/partial/no/NA)}
	NA

	\question{All novel datasets introduced in this paper will be made publicly available upon publication of the paper with a license that allows free usage for research purposes}{(yes/partial/no/NA)}
	NA

	\question{All datasets drawn from the existing literature (potentially including authors' own previously published work) are accompanied by appropriate citations}{(yes/no/NA)}
	NA

	\question{All datasets drawn from the existing literature (potentially including authors' own previously published work) are publicly available}{(yes/partial/no/NA)}
	NA

	\question{All datasets that are not publicly available are described in detail, with explanation why publicly available alternatives are not scientifically satisficing}{(yes/partial/no/NA)}
	NA

\end{itemize}
\end{itemize}

\checksubsection{Computational Experiments}
\begin{itemize}

\question{Does this paper include computational experiments?}{(yes/no)}
yes

\ifyespoints{If yes, please address the following points:}
\begin{itemize}

	\question{This paper states the number and range of values tried per (hyper-) parameter during development of the paper, along with the criterion used for selecting the final parameter setting}{(yes/partial/no/NA)}
	yes

	\question{Any code required for pre-processing data is included in the appendix}{(yes/partial/no)}
	NA

	\question{All source code required for conducting and analyzing the experiments is included in a code appendix}{(yes/partial/no)}
	yes

	\question{All source code required for conducting and analyzing the experiments will be made publicly available upon publication of the paper with a license that allows free usage for research purposes}{(yes/partial/no)}
	yes
        
	\question{All source code implementing new methods have comments detailing the implementation, with references to the paper where each step comes from}{(yes/partial/no)}
	yes

	\question{If an algorithm depends on randomness, then the method used for setting seeds is described in a way sufficient to allow replication of results}{(yes/partial/no/NA)}
	yes

	\question{This paper specifies the computing infrastructure used for running experiments (hardware and software), including GPU/CPU models; amount of memory; operating system; names and versions of relevant software libraries and frameworks}{(yes/partial/no)}
	yes

	\question{This paper formally describes evaluation metrics used and explains the motivation for choosing these metrics}{(yes/partial/no)}
	yes

	\question{This paper states the number of algorithm runs used to compute each reported result}{(yes/no)}
	yes

	\question{Analysis of experiments goes beyond single-dimensional summaries of performance (e.g., average; median) to include measures of variation, confidence, or other distributional information}{(yes/no)}
	yes

	\question{The significance of any improvement or decrease in performance is judged using appropriate statistical tests (e.g., Wilcoxon signed-rank)}{(yes/partial/no)}
	yes

	\question{This paper lists all final (hyper-)parameters used for each model/algorithm in the paper’s experiments}{(yes/partial/no/NA)}
	yes

\end{itemize}
\end{itemize}

\end{document}